\documentclass[11pt]{article}

\usepackage[preprint]{acl}

\usepackage{times}
\usepackage{latexsym}

\usepackage[T1]{fontenc}

\usepackage[utf8]{inputenc}

\usepackage{microtype}

\usepackage{inconsolata}

\usepackage{graphicx}

\usepackage[table]{xcolor}
\definecolor{mygreen}{RGB}{34,139,34} 
\usepackage{url}            
\usepackage{booktabs}       
\usepackage{amsfonts}       
\usepackage{nicefrac}       
\usepackage{tcolorbox}
\usepackage{amsmath}
\usepackage{array}
\usepackage{makecell}
\usepackage{longtable}
\usepackage{threeparttable}  
\usepackage{threeparttablex} 
\usepackage{caption}
\usepackage{pifont}
\usepackage{enumitem}
\usepackage{titletoc}
\usepackage{amssymb}
\usepackage{multirow}
\usepackage[normalem]{ulem}
\usepackage{tabularx}
\usepackage{xurl}
\usepackage{placeins} 

\definecolor{red}{RGB}{255,222,222} 
\definecolor{bred}{RGB}{230,57,70} 
\definecolor{purp}{RGB}{169, 57, 230} 

\title{Adapt or Calibrate?\\How Local Labels Buy Certified Coverage in Medical VQA}

\author{
\textbf{Reza Khanmohammadi}\thanks{Corresponding author} \\
Michigan State University \\
\texttt{khanreza@msu.edu}
\And
\textbf{Kundan Thind}\thanks{Shared senior authorship} \\
Henry Ford Health \\
\texttt{kthind1@hfhs.org}
\And
\textbf{Mohammad M. Ghassemi}\footnotemark[2] \\
Michigan State University \\
\texttt{ghassem3@msu.edu}
}

\begin{document}
\maketitle

\begin{abstract}
An answer from a vision-language model (VLM) to a medical image question can be accepted without review only under a threshold bounding the error of accepted answers, and certifying that bound costs local labels. We ask what such labels can certify for nine confidence estimators on five general-purpose open-weight VLMs and three medical VQA benchmarks under Learn-then-Test, with the question as the exchangeable unit. Estimators trained on natural images transfer poorly: the best pooled certificate covers 11.5\% of SLAKE answers at a 20\% error tolerance and nothing on PATH-VQA. With 1{,}000 labeled SLAKE questions, adapting a probe on half of them raises pooled certified coverage from 6.8\% to 37.6\% (0.0\% to 30.6\% with the image as the unit), above the best threshold-only estimator (11.3\%). On PATH-VQA, adaptation certifies 11.4\% with 2{,}500 labels; thresholds alone certify nothing. Yet for strong single models, small budgets are better spent on the threshold. A question-only classifier recovers much of adaptation's ranking gain but certifies far less. Neither procedure yields a pooled certificate at 10\%, and nothing is certified at 5\%. A blank image removes 61\% of correct answers, yet on GMAI-MMBench no trained probe is less confident on answers unchanged by image replacement.
\end{abstract}
\section{Introduction}
\label{sec:intro}

\noindent\textbf{A confidence score is useful for medical image question answering only if it can bound the error of the answers it lets through.} General-purpose open-weight vision-language models (VLMs) answer questions about radiographs, CT slices, and pathology slides, and sites that cannot send patient images to closed APIs can run them on their own hardware~\cite{umeton2024gpt4,wiest2025deidentifying}. Accuracy alone does not say which of their answers can be accepted without review, and that decision, not accuracy, determines how much review work a model can remove~\cite{sinsky2016allocation,ebbers2022quantifying}. It needs an acceptance threshold, fixed before use, under which the error rate among accepted answers stays below a stated tolerance on future questions, not only on the questions used to choose it~\cite{angelopoulos2021learn,angelopoulos}. Such a guarantee must be computed from labeled questions drawn from the site's own distribution, so it has a cost in labels. The practical question is what a given number of local labels can certify.

\noindent\textbf{Ranking and calibration metrics do not answer that question.} AUROC measures how well a score orders correct above incorrect answers, and expected calibration error measures how closely its values track accuracy, both on average. Neither determines how much of a workload can be accepted at a controlled error rate, which depends on the region of highest confidence and on how many labels are spent certifying it. Selective prediction makes this target explicit for visual question answering~\cite{whitehead,tejas} and for question answering under domain shift~\cite{amita}, and conformal methods provide finite-sample guarantees for vision-language outputs~\cite{zhuohang,asif,tayebati}. Recent studies of medical VQA document overconfidence and propose ways to reduce it~\cite{byun,yuetian,senoglu}. They report calibration and ranking metrics; none measures certified coverage, and labels enter only as the data needed to fit a recalibration map~\cite{byun}.

\noindent\textbf{A site starts from confidence estimators built on other data.} Trained confidence probes are usually fit on general-domain data, so a site without medical labels can only use them as transferred. We study nine estimators (training-free logit baselines, prompted self-reports, and internal-state probes trained on natural-image GQA~\cite{GQA}) on five general-purpose open-weight VLMs and three medical VQA benchmarks: SLAKE~\cite{SLAKE}, GMAI-MMBench~\cite{gmaimmbench}, and PATH-VQA~\cite{PATHVQA}. No estimator is best across datasets and models. With no labels at all, a site can only trust a fixed cut such as 0.8, which keeps the error among accepted answers below 20\% for some models and datasets and exceeds it for others, with no way to tell which in advance. When a labeled calibration set (about 1,000 questions on SLAKE) is used only to set the threshold of a transferred estimator, the best certificate pooled over the five models covers 11.5\% of SLAKE answers at a 20\% error tolerance, and no pooled certificate exists at 10\% on any dataset. Per model, the best transferred estimator certifies 47.8--54.6\% of SLAKE answers on each of the three most accurate backbones, so the pooled figure describes no single deployment.

\noindent\textbf{Once a site pays for labels, it must decide how to spend them.} The same labels can set the acceptance threshold of a transferred score, or be split between adapting the probe in domain and setting the threshold. Adaptation can improve the ranking, but splitting halves the labels left to certify the threshold, so the better use depends on the budget and the model. Pooled over the five models, adapting a probe never certifies less than calibrating the threshold of the same probe, for either adapted probe, at any budget we test. On SLAKE, at 1,000 labeled questions and a 20\% tolerance, spending half of them on adaptation raises the pooled certified coverage of the same probe from 6.8\% to 37.6\%, well above the 11.3\% of the best threshold-only estimator, mean token entropy. On PATH-VQA, threshold-only calibration certifies nothing at any budget, while adaptation reaches 11.4\% pooled coverage at 2,500 labeled questions. For a single model, threshold-only calibration can win when the transferred score is already strong, most often at small budgets: with 500 labeled SLAKE questions it certifies 48.7\% of one model's answers, against 5.1\% after adaptation. 

\noindent\textbf{A gain in certified coverage does not show what the adapted score learned.} We separate three properties that are easy to conflate: whether a score predicts correctness, whether the model used the image, and whether confidence registers when it did not. A classifier trained on the same labeled questions but given only the question text out-ranks the transferred probe on all three datasets and recovers much of the adapted probe's ranking gain, but it certifies less than half as much on SLAKE and nothing on GMAI-MMBench or PATH-VQA, and the stronger adapted probe adds ranking signal beyond it in every model and draw. Replacing the image with a blank one removes about 60\% of the models' correct answers, so their answers depend on the image in aggregate. However, on GMAI-MMBench answers that stay the same when the image is replaced by unrelated natural images, only P(True) lowers its confidence, and no trained probe does, before or after adaptation.

\noindent\textbf{Contributions.} We propose no new estimator or risk-control method; we analyze what existing ones can certify under a label budget. (i) An evaluation of nine transferred confidence estimators on five VLMs and three medical VQA benchmarks under Learn-then-Test risk control, with the question rather than each model answer as the exchangeable unit; under the same bound, treating a question's five correlated answers as independent would report 26.9\% certified coverage for one probe on SLAKE instead of the valid 6.0\%, at a threshold that fails to hold the tolerance. (ii) A matched-budget comparison of a fixed cut, threshold-only calibration, and adaptation plus calibration, showing that how local labels are allocated, and not only how many there are, decides certified coverage, and that the better allocation changes with the budget and the model. (iii) A question-text control showing that much of the adapted BICR probe's ranking gain is recoverable from the question and the transferred score, but most of its certified coverage is not, and image tests showing that the models use the image while no trained probe is less confident on answers that image replacement leaves unchanged.
\section{Related Work}
\label{sec:related}

\noindent\textbf{Selective prediction and risk control.} Selective prediction lets a model abstain and measures the coverage it keeps at a given risk. In visual question answering, \citet{whitehead} show that softmax scores support little coverage at low risk and that a learned selector raises it, and \citet{tejas} reduce unnecessary abstention by gathering further evidence from the image. Under domain shift, a calibrator trained on the model's own errors improves selective question answering~\cite{amita}; our adapted probes play a similar role, but we fix the label budget and ask how to divide it between fitting the score and certifying its threshold. Learn-then-Test and conformal risk control turn a labeled calibration set into finite-sample guarantees~\cite{angelopoulos2021learn,angelopoulos}. For vision-language models, conformal methods filter unsupported claims, including in radiology report generation~\cite{zhuohang}, benchmark uncertainty across many models~\cite{asif}, adapt abstention policies to context~\cite{tayebati}, and decide when to examine the image again before abstaining~\cite{Xu}; the last reports that keeping hallucination below 5\% on an object-existence benchmark requires abstaining on over 80\% of claims, illustrating the difficulty of achieving useful coverage at strict risk tolerances. These works treat the calibration set as given, whereas we vary its size and its use. Because five models answer each question, we take the question as the exchangeable unit, as in hierarchical conformal prediction~\cite{Dunn}.

\noindent\textbf{Confidence in medical VQA.} Trustworthiness benchmarks expose factual errors and reliability limitations in medical VLMs across diverse imaging modalities~\cite{xia}. \citet{byun} find on three medical VQA benchmarks, including SLAKE, that general-purpose VLMs remain overconfident across scales and prompting strategies, and that post-hoc recalibration lowers calibration error without improving discrimination; their hallucination-aware recalibration improves both and is fit on a small labeled set. \citet{yuetian} reduce the calibration error of verbalized confidence through multi-strategy interrogation with an auxiliary expert model, and \citet{senoglu} fine-tune a medical and a general-purpose VLM to verbalize better-calibrated uncertainty. These studies report calibration and ranking metrics. We instead measure how many answers can be accepted under a certified error tolerance and how that number depends on the way local labels are spent. Like \citet{byun}, we study general-purpose VLMs, and we adapt only auxiliary confidence estimators; the answering model is never changed.

\noindent\textbf{What a confidence score reflects.} Answers in visual question answering can often be predicted from the question alone, and benchmarks have been rebuilt to limit this shortcut~\cite{goyal2017making,agrawal2018dont}; access to the image likewise does not ensure that an answer depends on it~\cite{Woo2025AVISC}. A confidence score can exploit the same shortcut by learning which kinds of questions a model answers correctly, without regard to the image. Some estimators are built to be sensitive to the image, by separating visual from reasoning confidence~\cite{wenyi}, training on image-ablated inputs~\cite{senoglu}, or contrasting real-image and blank-image representations~\cite{bicr}, and internal-state uncertainty has been used to detect hallucinations without reference answers~\cite{li2024referencefree}. We separate question-type signal from image use directly: we compare adapted probes against a classifier that sees only the question text, and test whether any estimator assigns lower confidence to answers that stay the same when the image is replaced.
\section{Study Design}
\label{sec:setup}

{\looseness=-1
\noindent\textbf{Data and splits.} We use three medical VQA benchmarks. SLAKE~\cite{SLAKE} contains open and closed questions about CT, MRI, and X-ray images in English and Chinese (2{,}094 questions: 1{,}061 English, 1{,}033 Chinese). GMAI-MMBench~\cite{gmaimmbench} is a multiple-choice benchmark spanning more than forty clinical tasks across radiology, pathology, dermatology, ophthalmology, endoscopy, and surgical imaging (21{,}748 questions). PATH-VQA~\cite{PATHVQA} contains open and yes/no questions about pathology images (6{,}111 questions). We split each dataset once into a calibration half and a test half, by question and stratified by question category. All models share the split, and every reported result is measured on the same test half. The calibration half (1{,}047, 10{,}878, and 3{,}055 questions) is the pool of a site's local labels. Appendix~\ref{app:data} provides licensing and supplementary-data details.
\par}

\noindent\textbf{Models.} We evaluate five general-purpose open-weight VLMs: Qwen3-VL-8B~\cite{qwen3vl}, InternVL3.5-14B~\cite{internvl3_5}, Gemma-3-27B~\cite{gemma3}, LLaVA-NeXT-13B~\cite{llava}, and DeepSeek-VL2~\cite{deepseek-vl2}, all with greedy decoding and at most 64 new tokens. Mean accuracy across models is 60.9\% on SLAKE, 47.2\% on GMAI-MMBench, and 39.5\% on PATH-VQA (31--72\% per model; Table~\ref{tab:budget}).

\noindent\textbf{Correctness labels.} Free-form answers cannot be graded by string match: ``the cardiac silhouette is enlarged'' is a correct answer to a question whose gold answer is ``heart''. We therefore label each answer correct or incorrect ($y \in \{0,1\}$) with a \texttt{gpt-5-mini} judge that sees the image, the question, the gold answer, and the response, following common practice in confidence estimation~\cite{calibration-tuning,ccps,rmcb}. The judge's labels agree with a string matcher on 97.9\% of all closed items ($\kappa=0.958$) and with a blind non-expert annotator on 96.5\% of a stratified free-form sample ($\kappa=0.930$). Most residual disagreements are cases the judge marks wrong that the annotator credits, so the labels lean strict (Appendix~\ref{app:judge}).

\noindent\textbf{Confidence estimators.} We evaluate nine estimators in three families (Appendices~\ref{app:metrics} and~\ref{app:cost}). Two training-free logit baselines, the mean token log-likelihood (SeqLL) and the mean token entropy (PredEnt) of the generated answer, need no extra computation. Three prompt-based methods query the model again: P(True)~\cite{kadavath} reads the probability that the model judges its own answer correct, Self-Probing~\cite{self-probing} asks it to rate its answer from 0 to 100, and Prompt Ensemble~\cite{promptensemble} averages confidence over the question and ten paraphrases. Four trained probes read internal states. SAPLMA~\cite{saplma} applies an MLP to a hidden state, InternalInspector~\cite{internalinspector} applies a convolutional encoder to stacked internal states, CCPS~\cite{ccps} reads how representations respond to controlled perturbations, and BICR~\cite{bicr} is trained with a contrastive objective between representations of the real image and of a blank one. Each probe is trained per VLM on 20{,}000 natural-image GQA questions~\cite{GQA}, with hyperparameters selected on a further 5{,}000, using the architecture recommended by its source paper and averaging predictions over five seeds. Calibration is measured by 10-bin ECE, which we do not report for the logit baselines because their scores are not probabilities.

\noindent\textbf{How local labels are used.} We compare three regimes that a site could run with $n$ labeled questions from the calibration half.
\textbf{R0} uses no labels and accepts an answer when its confidence is at least $0.8$; we evaluate it per model for BICR and InternalInspector on their raw scores.
\textbf{R1} uses the $n$ labels only to certify an acceptance threshold for the transferred estimator.
\textbf{R2} splits the $n$ questions in half. One half trains a new probe from scratch for each VLM, with the same architecture and the hyperparameters already selected on GQA, so no labels are spent on hyperparameter search; a fifth of this half is held out for early stopping. The other half certifies its threshold. We adapt BICR and InternalInspector, the two trained probes with the highest transferred AUROC on every dataset. Budgets are $n \in \{100, 250, 500, 1000\}$ on SLAKE, and additionally $2{,}500$ on GMAI-MMBench and PATH-VQA. Section~\ref{sec:transfer} applies R1 to the whole calibration half. For each budget we draw five category-stratified samples without replacement, nested across budgets. A per-model result uses that model's answers to the $n$ questions and describes a single deployment; a pooled result uses all five models' answers and summarizes them under one guarantee. Two R2 variants, with different hyperparameters or probe head, appear in Appendix~\ref{app:deployment}.

\begin{table*}[t]
\centering
\small
\resizebox{\textwidth}{!}{%
\begin{tabular*}{\textwidth}{@{\extracolsep{\fill}}l rrr rrr rrr@{}}
\toprule
& \multicolumn{3}{c}{SLAKE} & \multicolumn{3}{c}{GMAI-MMBench} & \multicolumn{3}{c}{PATH-VQA} \\
\cmidrule(lr){2-4} \cmidrule(lr){5-7} \cmidrule(lr){8-10}
Estimator & AUROC & ECE\,\% & Cert.\,\% & AUROC & ECE\,\% & Cert.\,\% & AUROC & ECE\,\% & Cert.\,\% \\
\midrule
SeqLL & 0.659 & n/a & \textbf{11.5} & 0.626 & n/a & \textbf{7.1} & 0.651 & n/a & 0.0 \\
PredEnt & 0.658 & n/a & \textbf{11.5} & 0.625 & n/a & \textbf{7.1} & \textbf{0.653} & n/a & 0.0 \\
\midrule
P(True) & 0.646 & 24.0 & 9.9 & 0.617 & 32.4 & 0.0 & 0.561 & 38.5 & 0.0 \\
SelfProbe & 0.632 & 30.1 & 0.0 & 0.557 & 39.1 & 0.0 & 0.573 & 46.8 & 0.0 \\
PromptEns & 0.634 & 15.0 & 3.6 & 0.608 & 31.5 & 0.0 & 0.641 & 34.8 & 0.0 \\
\midrule
SAPLMA & 0.628 & 12.9 & 0.0 & 0.580 & 31.0 & 0.0 & 0.575 & 14.3 & 0.0 \\
II & 0.681 & 3.1 & 7.6 & 0.603 & 15.2 & 0.0 & 0.614 & 15.5 & 0.0 \\
CCPS & 0.543 & 17.7 & 0.0 & 0.523 & 29.4 & 0.0 & 0.556 & 24.3 & 0.0 \\
BICR & \textbf{0.695} & 20.3 & 6.0 & \textbf{0.627} & 11.3 & 2.1 & 0.589 & 14.9 & 0.0 \\
\bottomrule
\end{tabular*}%
}
\caption{Transferred confidence estimators, with no medical labels used for training. AUROC and 10-bin ECE are measured on the test half, pooled over the five VLMs. ECE is not reported for the two logit baselines, whose scores are not probabilities. Cert.\,\% is the pooled certified coverage at $\tau=0.20$: the share of test answers accepted by a threshold certified on the full calibration half with the question-unit betting bound ($\delta=0.05$). SeqLL and PredEnt are the mean token log-likelihood and mean token entropy; SelfProbe is Self-Probing; PromptEns is Prompt Ensemble; II is InternalInspector.}
\label{tab:transfer}
\end{table*}

\noindent\textbf{Certified selective prediction.} An answer is accepted when its confidence $c$ is at least a threshold $t$. Per model, each question contributes one answer, and we control the selective risk $R(t) = \Pr(y = 0 \mid c \ge t)$ with Learn-then-Test~\cite{angelopoulos2021learn}. We test 100 candidate thresholds at evenly spaced quantiles of the calibration confidences, each with a Bentkus upper confidence bound~\cite{bentkus2004} at level $\delta/100$ (Bonferroni, $\delta = 0.05$), require at least 30 accepted calibration answers, and keep the most permissive threshold whose risk is certified to be at most the tolerance $\tau$; building the grid instead from held-out unlabeled scores, so that it is independent of the certifying labels, lowers the largest-budget pooled certificates of the adapted probes by at most 1.3 points (Appendix~\ref{app:grid}). Pooled over models, a question's answers are correlated, so we take the question as the exchangeable unit. If $N_q(t)$ of question $q$'s answers are accepted and $A_q(t)$ of them are wrong, the pooled risk is the mean of $A_q(t)/N_q(t)$ over questions with $N_q(t) \ge 1$, bounded with a betting confidence bound~\cite{Waudby} under the same grid, level, and floor. This question-averaged risk is not the error rate over all accepted answers, but the realized answer-level error of every pooled certificate also stays below $\tau$ (Appendix~\ref{app:deployment:overshoot}). With probability at least $1-\delta$ over the calibration labels, the selected threshold keeps the risk at most $\tau$ on future questions exchangeable with the calibration questions. Our split assigns questions, not images, to the two halves, so the guarantee covers new questions about the benchmarks' images; SLAKE and PATH-VQA ask 11.6 and 7.1 questions per image, and for questions about unseen images the image is the appropriate unit (Appendix~\ref{app:deployment:unit}). The guarantee is relative to our correctness labels and says nothing about any individual answer. \emph{Coverage} is the fraction of test answers accepted, and \emph{certified coverage} is the coverage of the certified threshold, or zero when no threshold is certified. We report $\tau \in \{0.20, 0.10, 0.05\}$ as illustrative tolerances, not clinical ones. For comparison only, a \emph{plug-in} threshold is the most permissive one whose empirical error on the calibration labels is at most $\tau$; we never call it certified.
\section{Transferred Confidence}
\label{sec:transfer}

\noindent\textbf{Without medical training labels, no estimator is best, and ranking does not predict certification.} Table~\ref{tab:transfer} reports the nine estimators as a site would first receive them, with every threshold certified on the full calibration half. BICR has the highest AUROC on SLAKE and GMAI-MMBench, and mean token entropy has the highest on PATH-VQA, but the margins are small. Across the 15 model and dataset pairs, no estimator has the highest AUROC in more than a third of them, and none differs significantly from BICR after Holm correction (Appendix~\ref{app:metrics}). Neither ranking nor calibration determines how much can be certified. On GMAI-MMBench, BICR ranks best (AUROC 0.627) but certifies 2.1\% of answers, while mean token entropy (0.625) certifies 7.1\%. On SLAKE, InternalInspector has the lowest calibration error (3.1\%) but certifies 7.6\%, while the logit baselines, whose scores are not probabilities, certify 11.5\%. Certification depends on how cleanly a score separates its most confident answers, which neither AUROC nor ECE measures.

{\looseness=-1
\noindent\textbf{Pooled certificates are small; per-model certificates are not.} Pooled over the five models, the best transferred certificate at $\tau=0.20$ covers 11.5\% of SLAKE answers and 7.1\% of GMAI-MMBench answers, and no estimator certifies anything on PATH-VQA. Per model, the picture differs: on each of the three most accurate SLAKE backbones, the best of the four estimators in Table~\ref{tab:budget} certifies 47.8--54.6\% of answers. A pooled threshold must serve five models with accuracies from 31\% to 72\%, whereas a site deploying one model certifies it alone.
\par}

\noindent\textbf{Counting answers instead of questions overstates the guarantee.} Each SLAKE question is answered by all five models, so its answers are correlated. Under the same betting bound, treating every answer as an independent observation certifies 26.9\% of SLAKE answers for BICR, against 6.0\% with the question as the unit. The answer-level guarantee ignores this correlation and fails: on the test half, its threshold reaches a question-averaged risk of 20.9\%, against 9.1\% for the question-level one. The unit, not the choice of bound, drives the difference (Appendix Table~\ref{tab:supp:clustervalid}); with the image as the unit, no transferred estimator certifies anything on SLAKE.

\noindent\textbf{A threshold chosen without a bound does not hold its tolerance.} On the full calibration half, the plug-in threshold exceeds 20\% error on the test half in 14 of 36 per-model cells, and the certified threshold in 1 of 20 (Appendix~\ref{app:deployment:overshoot}). A fixed cut at 0.8, which needs no labels, exceeds it in 12 of the 28 per-model cells where it accepts any answer (Appendix~\ref{app:deployment:r0}).
\section{What Local Labels Buy}
\label{sec:budget}

\begin{figure*}[t]
\centering
\includegraphics[width=0.9\textwidth]{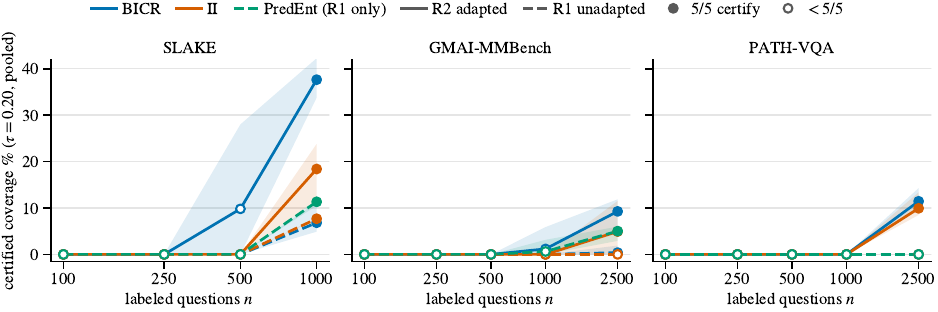}
\caption{Pooled certified coverage at $\tau=0.20$ against the number of labeled calibration questions $n$. Solid lines are R2, where half the labels train a new probe and half certify its threshold. Dashed lines are R1, where all labels certify the threshold of the transferred score; mean token entropy (PredEnt) has no adapted version. Coverage is the share of test answers accepted under the question-unit betting bound ($\delta=0.05$). Lines show the mean of five draws, bands the minimum and maximum, and filled markers the cells that certify in all five draws.}
\label{fig:budget}
\end{figure*}

\begin{table*}[t]
\centering
\small
\begin{tabular*}{\textwidth}{@{\extracolsep{\fill}}l l r l r r r r@{}}
\toprule
& & & & \multicolumn{4}{c}{Max budget, $\tau=0.20$} \\
\cmidrule(lr){5-8}
Dataset & LVLM & Acc. & Unadap. best & R1 BICR & R2 BICR & R1 II & R2 II \\
\midrule
\multirow{5}{*}{SLAKE} & InternVL-14B & 72.3 & 47.8 (P(True)) & 30.5 & 70.0 & 37.2 & 62.8 \\
 & Qwen3-VL-8B & 68.4 & 54.6 (II) & 54.1 & 64.9 & 54.2 & 60.1 \\
 & Gemma-27B & 64.6 & 50.7 (II) & 38.2 & 52.3 & 49.6 & 35.8$^{\dagger}$ \\
 & LLaVA-13B & 47.5 & 0 & 0.0 & 0.0 & 0.0 & 0.0 \\
 & DeepSeek-VL2 & 51.8 & 5.0 (II) & 4.4 & 12.4$^{\dagger}$ & 5.6 & 0.0 \\
\midrule
\multirow{5}{*}{GMAI} & InternVL-14B & 60.9 & 26.0 (P(True)) & 4.0$^{\dagger}$ & 10.8$^{\dagger}$ & 0.0 & 20.0 \\
 & Qwen3-VL-8B & 54.6 & 25.2 (PredEnt) & 4.8$^{\dagger}$ & 19.8 & 1.5$^{\dagger}$ & 19.6 \\
 & Gemma-27B & 48.9 & 6.3 (BICR) & 0.5$^{\dagger}$ & 2.5$^{\dagger}$ & 0.0 & 0.0 \\
 & LLaVA-13B & 35.1 & 0 & 0.0 & 0.0 & 0.0 & 0.0 \\
 & DeepSeek-VL2 & 37.0 & 0 & 0.0 & 0.0 & 0.0 & 0.0 \\
\midrule
\multirow{5}{*}{PATH-VQA} & InternVL-14B & 43.5 & 0 & 0.0 & 16.1 & 0.0 & 10.9 \\
 & Qwen3-VL-8B & 46.8 & 0 & 0.0 & 28.9 & 0.0 & 30.7 \\
 & Gemma-27B & 44.9 & 0 & 0.0 & 14.1$^{\dagger}$ & 0.0 & 8.1$^{\dagger}$ \\
 & LLaVA-13B & 30.8 & 0 & 0.0 & 0.0 & 0.0 & 2.8$^{\dagger}$ \\
 & DeepSeek-VL2 & 31.5 & 0 & 0.0 & 0.0 & 0.0 & 3.7$^{\dagger}$ \\
\bottomrule
\end{tabular*}
\caption{Per-model certified coverage (\%) at $\tau=0.20$ under Learn-then-Test with the Bentkus bound ($\delta=0.05$); coverage is the share of test answers accepted. Acc.\ is answer accuracy on all questions. Unadap.\ best is the best of BICR, InternalInspector (II), mean token entropy (PredEnt), and P(True), with the threshold certified on the full calibration half, or 0 if none certifies. R1 and R2 use the largest budget (1{,}000 questions on SLAKE, 2{,}500 on GMAI-MMBench and PATH-VQA) and are means over five draws. A dagger marks a nonzero mean in which at least one draw certified nothing.}
\label{tab:budget}
\end{table*}

\noindent\textbf{Adapting the probe buys pooled certificates that a threshold alone does not.} Figure~\ref{fig:budget} compares the two uses of the same labels. On SLAKE, with 1{,}000 labeled questions at $\tau=0.20$, R2 raises pooled certified coverage from 6.8\% to 37.6\% for BICR and from 7.7\% to 18.4\% for InternalInspector. The best threshold-only estimator, mean token entropy, reaches 11.3\%, and spending the adaptation half on per-model Platt scaling of the transferred score instead reaches 15.0\% for BICR (Appendix~\ref{app:deployment:recal}). On GMAI-MMBench, with 2{,}500 questions, adapted BICR certifies 9.3\% against 5.0\% for mean token entropy, while adapted InternalInspector does not exceed it (4.9\%). On PATH-VQA, no threshold-only estimator certifies anything at any budget, whereas adapted BICR and InternalInspector certify 11.4\% and 9.9\% at 2{,}500 questions, in all five draws. For the same probe, the pooled certificate after adaptation is never lower than with the threshold alone, at any budget. With the image as the unit, the SLAKE contrast at 1{,}000 questions becomes 0.0\% against 30.6\% for BICR, still certified in all five draws, while GMAI-MMBench and PATH-VQA change by at most 0.4 points (Appendix~\ref{app:deployment:unit}).

\noindent\textbf{Certificates require a minimum budget.} Pooled over models, nothing certifies at 250 labeled questions or fewer on any dataset. On SLAKE at 500 questions, adapted BICR certifies 9.8\%, but in only two of five draws. On GMAI-MMBench and PATH-VQA, certificates become reliable only at 2{,}500 questions, where adapted BICR certifies in every draw. Below these budgets, pooled certificates are absent or depend on the draw. Splitting the labels unevenly does not change the largest-budget results, but it shifts where certificates first appear: on SLAKE at 500 questions, adapting on a quarter of the labels and certifying on three quarters certifies 14.5\% in four of five draws (Appendix~\ref{app:deployment:split}).

\noindent\textbf{Per model, a threshold alone is sometimes the better use of the labels.} Table~\ref{tab:budget} reports the largest budget per model. At the largest budget, adaptation gives the largest certificates: 70.0\% for InternVL3.5-14B on SLAKE with BICR, 20.0\% for the same model on GMAI-MMBench with InternalInspector, and 30.7\% for Qwen3-VL-8B on PATH-VQA with InternalInspector. However, R2 certifies its threshold on half the labels, which costs most when the transferred score is already strong. On SLAKE, threshold-only calibration certifies more than adaptation in 1, 4, 3, and 2 of the 10 model and probe cells at 100, 250, 500, and 1{,}000 questions, while adaptation certifies more in 0, 1, 4, and 6. Nine of the ten cells where the threshold alone wins belong to Qwen3-VL-8B and Gemma-3-27B, the two models with the largest certificates before adaptation. With 500 labeled questions, Qwen3-VL-8B certifies 48.7\% of answers with the transferred BICR, against 5.1\% after adaptation; with 1{,}000, Gemma-3-27B certifies 49.6\% with the transferred InternalInspector, against 35.8\% after adaptation. The better allocation therefore depends on both the budget and the model.

\noindent\textbf{Certified thresholds hold their tolerance, and plug-in thresholds do not.} Across all budgets, draws, and models, the plug-in threshold exceeds 20\% error on the test half in 409 of the 778 cases where it yields an operating point (52.6\%), and the certified threshold in 1 of 210, at 20.3\%. At $\tau=0.10$ the counts are 350 of 514 and 0 of 34. The pooled question-unit certificate never exceeds its tolerance (0 of 54).

\noindent\textbf{Strict tolerances remain out of reach.} At $\tau=0.10$, no pooled certificate exists in any of 77 cells, with or without adaptation, and at $\tau=0.05$ nothing is certified in any of 462 cells; among uneven splits, a single draw certifies at $\tau=0.10$ (Appendix~\ref{app:deployment:split}). Single models certify at 10\% mostly on SLAKE: up to 26.9\% for InternVL3.5-14B after adaptation, and 21.6\% for Qwen3-VL-8B with the threshold alone, which beats adaptation there (Appendix Table~\ref{tab:tau10}).

\begin{table*}[t]
\centering
\small
\small
\begin{tabular*}{\textwidth}{@{\extracolsep{\fill}}l rrrr c rrrr@{}}
\toprule
& \multicolumn{4}{c}{AUROC} & Incr. & \multicolumn{4}{c}{Cert. \%} \\
\cmidrule(lr){2-5} \cmidrule(lr){7-10}
Dataset & Text & Text+R1 & R1 & R2 & R2 over Text & Text & Text+R1 & R1 & R2 \\
\midrule
SLAKE & 0.786 & 0.813 & 0.695 & 0.835 & $+0.048$ (25/25) & 13.3$^{\dagger}$ & 29.2 & 6.8 & 37.6 \\
GMAI-MMBench & 0.688 & 0.705 & 0.627 & 0.723 & $+0.034$ (25/25) & 0.0 & 0.0 & 0.4$^{\dagger}$ & 9.3 \\
PATH-VQA & 0.782 & 0.795 & 0.589 & 0.831 & $+0.058$ (25/25) & 0.0 & 1.0$^{\dagger}$ & 0.0 & 11.4 \\
\bottomrule
\end{tabular*}
\caption{Question-text control for BICR at the largest budget, $\tau=0.20$. Text sees only the question; Text+R1 adds the transferred probe's score to Text's features; R1 is the transferred probe and R2 the adapted one. AUROC is on the test half, pooled over models and draws. Incr.\ is the cross-fitted AUROC gain from adding the R2 score to Text, with the number of 25 model and draw tests whose 95\% interval is above zero. Cert.\ is pooled certified coverage as in Figure~\ref{fig:budget}, averaged over five draws; Text and R2 share the same label halves. A dagger marks a nonzero mean with at least one draw certifying nothing. InternalInspector: Appendix Table~\ref{tab:textcontrol_ii}.}
\label{tab:textcontrol}
\end{table*}
\section{What Adaptation Learns}
\label{sec:learns}

\noindent\textbf{Three controls separate what a score predicts from what the model looked at.} A gain in certified coverage could come from learning which questions a model tends to answer correctly, regardless of the image. We therefore train a \emph{question-text classifier}, a logistic regression on TF-IDF features of the question and its benchmark metadata, which never sees the image, the answer, or internal states. It uses the same adaptation and calibration halves as R2. We measure what the adapted probe adds beyond it by cross-fitting the classifier with and without the probe score on the test half (Appendix~\ref{app:permodel:text}). To test image use, we replace each image with a black one and regrade the answers. To test whether confidence registers image use, we regenerate answers to 5{,}000 GMAI-MMBench questions with the image replaced by three unrelated natural images, and call an answer \emph{image-invariant} if its text never changes. For each estimator we take the mean confidence on the other answers minus that on image-invariant answers, within each model, averaged over the four models with at least 30 image-invariant answers. Image-invariant answers are less accurate, so a calibrated score should drop by the accuracy gap, the \emph{warranted} drop.

\noindent\textbf{Question text explains much of the ranking, but the certificates of adapted BICR come from adaptation.} The text classifier out-ranks the transferred BICR on all three datasets (Table~\ref{tab:textcontrol}): AUROC 0.786 against 0.695 on SLAKE, 0.688 against 0.627 on GMAI-MMBench, and 0.782 against 0.589 on PATH-VQA. On SLAKE it certifies 13.3\%, against 6.8\% for the transferred probe. The adapted probe still adds signal beyond the text, by 0.048, 0.034, and 0.058 AUROC, with an interval above zero in all 25 model and draw tests on each dataset; these tests share test questions, so they show consistency rather than independent replication. Adding the transferred BICR score to the text classifier brings its AUROC to within 0.04 of the adapted probe, yet it certifies 29.2\%, 0.0\%, and 1.0\%, against 37.6\%, 9.3\%, and 11.4\% after adaptation. Ranking is nearly recoverable without adaptation; the certifiable region is not. For InternalInspector, adaptation helps less: the same combination out-ranks the adapted probe on SLAKE and GMAI-MMBench and certifies more on SLAKE (24.1\% against 18.4\%; Appendix Table~\ref{tab:textcontrol_ii}).

\noindent\textbf{The models use the image.} With a blank image, the models lose 60.9\% of the answers they got right with the real image (57.2--68.3\% per dataset), and 82.4\% on open-ended questions. On GMAI-MMBench, whose multiple-choice format gives a measurable chance level, blank-image accuracy is 26.0\% against a chance floor of 23.8\%.

\noindent\textbf{No trained probe lowers its confidence on image-invariant answers, before or after adaptation.} On GMAI-MMBench, image-invariant answers are less accurate, and the warranted drop is 1.71 points. Only P(True) lowers its confidence on them significantly, by 1.36 points (95\% interval 0.62 to 2.09). The four transferred probes either show no significant change (BICR $-0.72$, CCPS $-0.78$) or are significantly \emph{more} confident on image-invariant answers (SAPLMA $-2.04$, InternalInspector $-2.02$), as are the two logit baselines and the other prompt-based methods. Adaptation does not change this: averaged over draws, the drop is $-1.40$ for adapted BICR and $-0.37$ for adapted InternalInspector, and no draw is significantly positive.
\section{Discussion}
\label{sec:discussion}

\noindent\textbf{A certificate is a statistical statement, not a clinical one.} The guarantee bounds the average error among accepted answers on future questions drawn like the calibration questions, measured against our correctness labels. It does not say that any accepted answer is correct, that the labels match a clinician's judgment, or that a 20\% tolerance is acceptable for any clinical use. It also depends on exchangeability. If the questions a site receives differ from those it labeled, for example after a change of scanner, case mix, or question style, or for images not represented in calibration, the certificate no longer applies, and the site must label and certify again.

\noindent\textbf{How a site should spend its labels depends on the budget and the model.} Without labels, no threshold can be trusted. With 250 or fewer labeled questions, no pooled certificate exists, and per-model certificates come mostly from the threshold alone, on models whose transferred score is already strong. With about 1{,}000 labeled questions on SLAKE, or 2{,}500 on GMAI-MMBench and PATH-VQA, adapting the probe on half of them gives the largest pooled certificates, and at smaller budgets a smaller adaptation share certifies earlier. These budgets will differ elsewhere, but R1 and R2 use the same labels, so a site can compare both.

\noindent\textbf{Claims about adapted confidence should be checked against the question text.} A classifier that never sees the image out-ranks every transferred probe here and, combined with the transferred score, beats adapted InternalInspector on SLAKE. Reports that adaptation improves a confidence score should include such a baseline and measure certified coverage, not only AUROC. A score built to be sensitive to the image need not register when the model ignores it.
\section{Conclusion}
\label{sec:conclusion}

{\looseness=-1
We asked what local labels can certify for general-purpose VLMs answering medical image questions. With the question, not the model answer, as the unit, transferred confidence estimators certify little: pooled over models, at most 11.5\% of SLAKE answers at a 20\% tolerance, and nothing on PATH-VQA. Adapting the probe on half the labels gives the largest pooled certificates at 1{,}000 to 2{,}500 questions, while for a strong model at small budgets the threshold alone is better. Question text recovers much of what adaptation adds to ranking but little of its certified coverage, and no trained probe is less confident on answers that image replacement leaves unchanged. Certified selective prediction can reduce review at loose tolerances, but not yet at strict ones.
\par}
\section*{Limitations}
We study general-purpose open-weight VLMs only. Medical models such as MedGemma or LLaVA-Med are out of scope, and since certificates are specific to a model and its confidence scores, they would need to be certified separately. Correctness labels come from an LLM judge, validated against a string matcher, a blind non-expert annotator, and two other LLM judges, but not adjudicated by clinicians. All certificates hold relative to these labels, and the judge's residual errors, mostly rejections of answers the annotator credits, bias certified coverage downward. The questions come from VQA benchmarks, not a clinical workflow, so certified coverage describes these question distributions, not a site's workload, and the tolerances of 20\%, 10\%, and 5\% are illustrative, not clinically derived. The SLAKE calibration half has 1{,}047 questions, so its draws of 1{,}000 overlap heavily (mean pairwise Jaccard 0.91), as do PATH-VQA draws of 2{,}500 (0.69). The spread across draws at these budgets therefore understates the spread across independent label sets. Our split assigns questions rather than images to the two halves, so every SLAKE test image also appears in calibration, and the question-unit certificates cover new questions about these images rather than new images. With the image as the unit, SLAKE's transferred certificates vanish, while adapted BICR at 1{,}000 questions still certifies 30.6\% in all five draws (Appendix~\ref{app:deployment:unit}); an image-disjoint split would measure out-of-image performance directly. We adapt two probes with hyperparameters fixed from GQA, search hyperparameters on local labels only for BICR on SLAKE (Appendix~\ref{app:deployment:variants}), and test only three splits of the labels, without the 75/25 split on GMAI-MMBench (Appendix~\ref{app:deployment:split}). Finer splits could move where adaptation overtakes the threshold alone. The candidate thresholds are quantiles of the calibration confidences; building them from held-out unlabeled scores removes this dependence at a small cost in coverage (Appendix~\ref{app:grid}). The image-replacement test uses natural rather than medical images, defines image invariance by an unchanged answer, is informative only on GMAI-MMBench, and excludes one model with too few invariant answers. Finally, all errors count equally; a harm-weighted sensitivity analysis (Appendix~\ref{app:harm}) uses tiers drafted with an LLM and reviewed by the authors, not validated by clinicians.
\section*{Ethical Considerations}

\paragraph{A certificate can be over-trusted.} A certified threshold bounds question-averaged error at a chosen tolerance only when calibration and use questions are exchangeable; it is a statistical guarantee, not clinical validation, and it does not hold under shift between the two. The tolerances we study are illustrative, pooled certificates are essentially absent at a 10\% tolerance and absent at 5\%, and all models are general-purpose open-weight VLMs evaluated on public benchmarks rather than clinical systems.

\paragraph{AI assistants were used for code and writing.} We used AI coding assistants to write and refactor experiment and analysis code, AI assistants to help draft and edit the manuscript, and an LLM to draft the harm tiers of Appendix~\ref{app:harm}, which the authors reviewed. All code was verified by re-running results against stored outputs, and all numbers and claims were checked by the authors, who take full responsibility for the content. LLMs used as correctness judges and confidence estimators are research objects and are described in Section~\ref{sec:setup} and Appendix~\ref{app:judge}.
\bibliography{references}

\clearpage
\appendix
\onecolumn

\begin{center}
\huge\textbf{Appendix}
\end{center}

\begin{center}
\Large\textbf{Table of Contents}
\end{center}

\noindent\rule{\linewidth}{0.6pt}

{\hypersetup{linkcolor=black}
\let\OldSection\section
\renewcommand{\section}{\FloatBarrier\OldSection}
\startcontents[sections]
\printcontents[sections]{}{1}{}}

\noindent\rule{\linewidth}{0.6pt}

\twocolumn

\section{Metrics and Estimator Details}
\label{app:metrics}

\noindent\textbf{Notation.} Each test answer $i$ has a correctness label $y_i \in \{0,1\}$ and a confidence $s_i$, with larger $s_i$ meaning more likely correct. A per-model cell contains one answer per question. A pooled cell contains all five models' answers, so a question $q$ contributes up to five answers.

\noindent\textbf{AUROC and ECE.} AUROC is the probability that a randomly chosen correct answer receives a higher confidence than a randomly chosen incorrect one, with ties counted as one half. ECE partitions $(0,1]$ into 10 equal-width bins $B_1,\dots,B_{10}$, half-open on the left, and computes
\[
\mathrm{ECE} = \sum_{m=1}^{10} \frac{|B_m|}{N}\,\big|\mathrm{acc}(B_m) - \mathrm{conf}(B_m)\big|,
\]
where $\mathrm{acc}(B_m)$ is the fraction of correct answers in bin $m$ and $\mathrm{conf}(B_m)$ the mean confidence there. A confidence of exactly 0 falls in no bin; this affects only Self-Probing and CCPS, which can output 0. We do not report ECE for the two logit baselines, whose scores are not probabilities.

\noindent\textbf{Coverage and risk.} A threshold $t$ accepts the answers with $s_i \ge t$. Coverage is the fraction of test answers accepted, in both per-model and pooled cells. Per model, the selective risk is the error rate among accepted answers. Pooled, if $N_q(t)$ of question $q$'s answers are accepted and $A_q(t)$ of them are wrong, the risk is
\begin{equation}
\begin{aligned}
R(t) &= \frac{1}{|Q(t)|}
\sum_{q \in Q(t)} \frac{A_q(t)}{N_q(t)},\\
Q(t) &= \{q : N_q(t) \ge 1\}.
\end{aligned}
\end{equation}
the error rate among a question's accepted answers, averaged over questions with at least one accepted answer. Questions with no accepted answer do not enter the average. Where we also report the error rate over all accepted answers, we call it the answer-level error.

\noindent\textbf{Thresholds.} Every candidate threshold is a confidence value observed on the calibration labels, so a group of tied confidences is always accepted or rejected as a whole. Every rule requires at least 30 accepted calibration answers. The certified rule is described in Section~\ref{sec:setup} and Appendix~\ref{app:deployment}. The plug-in rule searches every distinct calibration confidence, not the 100-point grid, and picks the most permissive threshold whose empirical calibration risk is at most $\tau$. It carries no guarantee. The oracle yield $Y@\tau$ is the largest coverage whose empirical risk on the evaluated data is at most $\tau$, with the same 30-answer floor. Because it chooses the threshold on the data it scores, we use it only in the harm-weighted analysis of Appendix~\ref{app:harm}.

\noindent\textbf{Logit baselines.} Both are computed from the top 100 next-token logits stored at each generated position, renormalized over those 100 tokens. Mean token log-likelihood is $\exp\big(\frac{1}{T}\sum_{t=1}^{T} \log p_t\big)$, where $p_t$ is the probability of the generated token at position $t$ and $T$ the answer length. Mean token entropy is $1 - \bar H / \log 100$, where $\bar H$ is the entropy of the renormalized distribution averaged over the $T$ positions.

\noindent\textbf{Prompt-based estimators.} P(True) appends the question ``Is the proposed answer correct? A) no B) yes'' and reads the probabilities of the two option tokens at the first generated position, which costs one extra forward pass. Self-Probing reconstructs the conversation, asks the model to rate its answer from 0 to 100, and parses the integer. Prompt Ensemble generates answers to ten paraphrases of the question and averages the confidence over the original and the ten paraphrases.

\noindent\textbf{Trained probes.} SAPLMA applies an MLP to one hidden-state vector. InternalInspector applies a residual convolutional encoder to a three-channel stack of internal states. CCPS applies a convolutional network to the model's internal states under a sequence of controlled perturbations. BICR applies an MLP to the last-layer hidden state and is trained with a ranking term that contrasts the real image with a blank one. Each probe is trained per VLM on 20{,}000 GQA questions. Hyperparameters and early stopping are selected on 5{,}000 further GQA questions by $0.6\,\mathrm{AUROC} + 0.4\,(1-\mathrm{ECE})$, and predictions are averaged over the seeds 23, 42, 137, 2024, and 3407. For R2, the same architecture and GQA-selected hyperparameters are trained from scratch on the adaptation half, with 20\% of that half held out for early stopping on the same criterion (patience 20, at most 200 epochs).

\noindent\textbf{Significance tests.} For the comparison in Section~\ref{sec:transfer}, we compute each estimator's AUROC on the test half in each of the 15 model and dataset cells, apply a Friedman test across the nine estimators, and compare each estimator with BICR by a Wilcoxon signed-rank test over the 15 cells, with Holm correction over the eight comparisons. The Friedman test is significant ($\chi^2=19.56$, $p=0.012$), but no Wilcoxon comparison with BICR survives Holm correction (smallest adjusted $p=0.38$).

\clearpage
\section{Certification and Label Budget}
\label{app:deployment}

\subsection{Procedure}
\label{app:deployment:procedure}

\noindent\textbf{Per model.} For $n$ calibration answers with confidences sorted in decreasing order, the candidate thresholds are the confidences at the 100 ranks $\mathrm{round}(1), \dots, \mathrm{round}(n)$ evenly spaced between 1 and $n$. For each candidate with at least 30 accepted answers, we compute the Bentkus upper confidence bound on the error rate among accepted answers. For a 0/1 loss this equals the one-sided Clopper--Pearson bound at level $\delta'/e$, where $\delta' = \delta/100$ and $\delta = 0.05$. The certified threshold is the most permissive candidate whose bound is at most $\tau$. Because every candidate is tested at level $\delta/100$, the family of tests holds jointly with probability at least $1-\delta$ (Bonferroni), and no fixed testing order is needed.

\noindent\textbf{Pooled.} The candidates are built in the same way from the pooled calibration confidences of the five models. The loss of a question is $A_q(t)/N_q(t)$, the error rate among its accepted answers, over questions with at least one accepted answer (Appendix~\ref{app:metrics}). We bound its mean with the hedged betting confidence bound of Waudby-Smith and Ramdas, with hedging parameter $1/2$ and bets truncated at $1/2$, evaluated on a grid of 1{,}001 candidate means, at level $\delta/100$. The floor of 30 counts accepted answers.

\noindent\textbf{Plug-in.} The plug-in threshold is the most permissive distinct calibration confidence whose empirical risk is at most $\tau$ with at least 30 accepted answers. It searches more candidates than the certified rule and carries no guarantee.

\subsection{The exchangeable unit}
\label{app:deployment:unit}

\begin{table*}[t]
\centering
\small
\resizebox{\textwidth}{!}{%
\begin{tabular}{@{}l@{\hspace{6pt}}r@{\hspace{4pt}}r@{\hspace{4pt}}r@{\hspace{4pt}}r@{\hspace{7pt}}r@{\hspace{4pt}}r@{\hspace{4pt}}r@{\hspace{7pt}}r@{\hspace{4pt}}r@{\hspace{4pt}}r@{\hspace{7pt}}l@{}}
\toprule
 & \multicolumn{4}{c}{Accepted \emph{row}} & \multicolumn{3}{c}{Accepted \emph{question}} & \multicolumn{3}{c}{Accepted \emph{image}} & \\
\cmidrule(lr){2-5}\cmidrule(lr){6-8}\cmidrule(lr){9-11}
Cell & Bentkus & Hoeff. & E.-Bern. & WSR & Hoeff. & E.-Bern. & WSR & Hoeff. & E.-Bern. & WSR & WSR err.\ \% [95\% CI] \\
\midrule
BICR, SLAKE & 24.5 & 19.5 & 19.5 & 26.9 & 0.0 & 0.0 & 6.0 & 0.0 & 0.0 & 0.0 & 7.03 [4.26, 10.07] \\
PredEnt$^{*}$, GMAI & 9.2 & 8.1 & 8.1 & 9.2 & 6.0 & 7.1 & 7.1 & 6.0 & 6.0 & 7.1 & 14.90 [13.65, 16.07] \\
\bottomrule
\end{tabular}
}
\caption{Pooled certified coverage (\%) at $\tau=0.20$ on the full calibration half, varying only the exchangeable unit and the concentration bound. Every column uses the same 100-point threshold grid, Bonferroni level $\delta/100$ with $\delta=0.05$, and 30-answer floor, and coverage is the share of test answers accepted. The \emph{row} unit treats each model answer as independent; the \emph{question} unit bounds the mean, over questions, of the error rate among a question's accepted answers, and the \emph{image} unit the same mean over images. The bounds are the binomial tail bound of Bentkus, Hoeffding, empirical Bernstein, and the betting bound of Waudby-Smith and Ramdas (WSR). WSR err.\ is the answer-level error on the test half at the question-unit WSR threshold, with a question-clustered bootstrap interval. PredEnt$^{*}$ is the mean token entropy, a training-free logit method.}
\label{tab:supp:clustervalid}
\end{table*}

Table~\ref{tab:supp:clustervalid} varies the unit and the bound with everything else fixed. With the betting bound, treating each answer as independent certifies 26.9\% of SLAKE answers for BICR, against 6.0\% with the question as the unit. The row-unit threshold does not hold the question-averaged risk it is meant to control: on the test half that risk is 20.88\% for BICR on SLAKE and 20.02\% for mean token entropy on GMAI-MMBench. The question-unit thresholds hold at 9.13\% and 16.81\%. Within either unit, the choice among the bounds changes coverage by much less than the change of unit.

The question unit is itself not the coarsest dependence in the data. SLAKE's 2{,}094 questions concern 180 images (11.6 per image) and PATH-VQA's 6{,}111 concern 858 (7.1); GMAI-MMBench has one question per image. Because the split randomizes questions, every SLAKE image and 97\% of PATH-VQA test questions' images also appear in calibration, so the question-unit certificate holds for new questions about these images. For questions about unseen images, we repeat the pooled rule with the image as the unit, the loss being the error rate among an image's accepted answers (Tables~\ref{tab:supp:clustervalid} and~\ref{tab:imageunit}). On SLAKE, with 180 rather than 1{,}047 units, no transferred estimator certifies anything on the full calibration half; at every budget, threshold-only BICR and InternalInspector certify nothing, and mean token entropy certifies 1.4\% at 1{,}000 questions in one draw of five. Adapted BICR at 1{,}000 questions still certifies 30.6\% in all five draws, against 37.6\% with the question unit, and adapted InternalInspector 7.8\% in three. The failure is the bound's width, not the floor: for transferred BICR, the tightest bound on the grid is 0.221. The question-text control loses more than the adapted probe: with the transferred BICR score, it falls from 29.2\% to 16.3\% on SLAKE, so adapted BICR's lead over it grows from 8.5 to 14.3 points. On GMAI-MMBench and PATH-VQA, every pooled certificate changes by at most 0.4 points. No certified threshold under either unit exceeds its tolerance on the test half, or on the test questions whose image is absent from calibration (GMAI-MMBench and 3\% of PATH-VQA; none of SLAKE). Per model, the image unit narrows adaptation's advantage on SLAKE: with BICR at 1{,}000 questions, adaptation certifies more in 13 of 25 model and draw pairs rather than 18.

\begin{table*}[t]
\centering
\small
\small
\begin{tabular}{@{}ll@{\hspace{10pt}}r@{\hspace{10pt}}r@{}}
\toprule
Dataset & Arm & Question unit & Image unit \\
\midrule
SLAKE & Best transferred, full calib. & 11.5 & 0.0 \\
 & R1 BICR & 6.8 & 0.0 \\
 & R2 BICR & 37.6 & 30.6 \\
 & R1 II & 7.7 & 0.0 \\
 & R2 II & 18.4 & 7.8$^{\dagger}$ \\
\midrule
GMAI-MMBench & Best transferred, full calib. & 7.1 & 7.1 \\
 & R1 BICR & 0.4$^{\dagger}$ & 0.0 \\
 & R2 BICR & 9.3 & 9.7 \\
 & R1 II & 0.0 & 0.0 \\
 & R2 II & 4.9$^{\dagger}$ & 5.1$^{\dagger}$ \\
\midrule
PATH-VQA & Best transferred, full calib. & 0.0 & 0.0 \\
 & R1 BICR & 0.0 & 0.0 \\
 & R2 BICR & 11.4 & 11.1 \\
 & R1 II & 0.0 & 0.0 \\
 & R2 II & 9.9 & 10.2 \\
\bottomrule
\end{tabular}

\caption{Pooled certified coverage (\%) at $\tau=0.20$ with the question or the image as the exchangeable unit, on the full calibration half (best transferred estimator) and at the largest budget (mean of five draws). A dagger marks a nonzero mean in which at least one draw certified nothing.}
\label{tab:imageunit}
\end{table*}

\subsection{Sensitivity to the threshold grid}
\label{app:grid}

\begin{table*}[t]
\centering
\small
\small
\begin{tabular}{@{}l rrr@{\hspace{5pt}} rrr@{\hspace{5pt}} rrr@{}}
\toprule
& \multicolumn{3}{c}{SLAKE} & \multicolumn{3}{c}{GMAI-MMBench} & \multicolumn{3}{c}{PATH-VQA} \\
\cmidrule(lr){2-4}\cmidrule(lr){5-7}\cmidrule(lr){8-10}
Estimator & Calib. & 10\% & 20\% & Calib. & 10\% & 20\% & Calib. & 10\% & 20\% \\
\midrule
\multicolumn{10}{@{}l}{\textit{Full calibration half, pooled, $\tau=0.20$}} \\
SeqLL & 11.5 & 11.1 & 10.8 & 7.1 & 6.8 & 6.2 & 0.0 & 0.0 & 0.0 \\
PromptEns & 3.6 & 0.0 & 0.0 & 0.0 & 0.0 & 0.0 & 0.0 & 0.0 & 0.0 \\
BICR & 6.0 & 7.7 & 6.5 & 2.1 & 1.9 & 2.1 & 0.0 & 0.0 & 0.0 \\
II & 7.6 & 7.5 & 6.5 & 0.0 & 0.0 & 0.0 & 0.0 & 0.0 & 0.0 \\
PredEnt & 11.5 & 10.8 & 8.5 & 7.1 & 6.7 & 7.3 & 0.0 & 0.0 & 0.0 \\
P(True) & 9.9 & 9.9 & 0.0 & 0.0 & 0.0 & 0.0 & 0.0 & 0.0 & 0.0 \\
SelfProbe & 0.0 & 0.0 & 0.0 & 0.0 & 0.0 & 0.0 & 0.0 & 0.0 & 0.0 \\
SAPLMA & 0.0 & 0.0 & 0.0 & 0.0 & 0.0 & 0.0 & 0.0 & 0.0 & 0.0 \\
CCPS & 0.0 & 0.0 & 0.0 & 0.0 & 0.0 & 0.0 & 0.0 & 0.0 & 0.0 \\
\midrule
\multicolumn{10}{@{}l}{\textit{Label budget, largest $n$ per dataset, pooled, $\tau=0.20$, mean of 5 draws}} \\
R1 BICR & 6.8 & 2.7 & 0.5 & 0.4 & 0.0 & 0.0 & 0.0 & 0.0 & 0.0 \\
R1 II & 7.7 & 7.5 & 6.4 & 0.0 & 0.0 & 0.0 & 0.0 & 0.0 & 0.0 \\
R1 PredEnt & 11.3 & 11.3 & 10.4 & 5.0 & 4.8 & 4.5 & 0.0 & 0.0 & 0.0 \\
R2 BICR & 37.6 & 36.9 & 35.3 & 9.3 & 8.1 & 7.5 & 11.4 & 10.2 & 9.7 \\
R2 II & 18.4 & 17.4 & 17.0 & 4.9 & 4.6 & 2.9 & 9.9 & 9.8 & 9.8 \\
\bottomrule
\end{tabular}

\caption{Pooled certified coverage (\%) at $\tau=0.20$ with candidate thresholds built from the certifying calibration scores (Calib.) or from a held-out slice of 10\% or 20\% of the calibration questions, whose labels are never used. Upper block: full calibration half. Lower block: largest budget (1{,}000 SLAKE questions, 2{,}500 GMAI-MMBench and PATH-VQA questions), mean of five draws.}
\label{tab:heldout}
\end{table*}

The candidate thresholds are quantiles of the calibration confidences, so the grid depends on the data that certify it, while Learn-then-Test assumes the candidates are fixed in advance. We therefore repeat every certification with a grid built from a held-out slice: a random 10\% or 20\% of the certification questions contributes only its confidences, never its labels, to the 100 candidate quantiles, and the remaining questions certify. This makes the candidates independent of the certifying labels at the cost of a smaller certification set. A site could avoid that cost by building the grid from unlabeled questions. With a 10\% slice, the pooled certificates of the adapted probes at the largest budgets fall by at most 1.25 points (BICR 36.9\%, 8.1\%, and 10.2\%), and every draw still certifies (Table~\ref{tab:heldout}). Small transferred certificates are less stable: on SLAKE, Prompt Ensemble loses its 3.6\% certificate, and threshold-only BICR at 1{,}000 questions falls from 6.8\% to 2.7\%, certifying in three of five draws. Per model, on SLAKE, threshold-only calibration still certifies more than adaptation in 0, 3, 3, and 2 cells at 100 to 1{,}000 questions. No pooled certificate exists at $\tau=0.10$ or $\tau=0.05$ under either grid, and no pooled certificate exceeds its tolerance on the test half.

\subsection{Certified and plug-in thresholds}
\label{app:deployment:overshoot}

\begin{table*}[t]
\centering
\small
\begin{tabular}{@{}l@{\hspace{4pt}}l@{\hspace{6pt}}r@{\hspace{3pt}}r@{\hspace{5pt}}r@{\hspace{6pt}}r@{\hspace{3pt}}r@{\hspace{5pt}}r@{\hspace{6pt}}r@{\hspace{3pt}}r@{\hspace{5pt}}r@{\hspace{6pt}}r@{\hspace{3pt}}r@{\hspace{5pt}}r@{}}
\toprule
& & \multicolumn{3}{c}{BICR} & \multicolumn{3}{c}{II} & \multicolumn{3}{c}{PredEnt} & \multicolumn{3}{c}{P(True)} \\
\cmidrule(lr){3-5} \cmidrule(lr){6-8} \cmidrule(lr){9-11} \cmidrule(lr){12-14}
Dataset & LVLM & Cov. & Err. & Cert. & Cov. & Err. & Cert. & Cov. & Err. & Cert. & Cov. & Err. & Cert. \\
\midrule
\multirow{5}{*}{SLAKE} & InternVL-14B & 54.7 & \textbf{21.8} & 32.2 & 66.2 & 18.8 & 36.6 & 63.2 & 19.5 & 45.2 & 65.3 & 19.4 & 47.8 \\
 & Qwen3-VL-8B & 68.2 & 18.2 & 53.8 & 72.1 & 19.2 & 54.6 & 35.2 & 15.7 & 0.0 & 46.7 & 17.4 & 33.5 \\
 & Gemma-27B & 55.1 & \textbf{21.7} & 38.1 & 63.3 & \textbf{24.4} & 50.7 & 35.9 & 19.7 & 24.4 & 45.6 & \textbf{23.9} & 28.4 \\
 & LLaVA-13B & 3.6 & \textbf{21.1} & 0.0 & n/a & n/a & 0.0 & n/a & n/a & 0.0 & n/a & n/a & 0.0 \\
 & DeepSeek-VL2 & 23.0 & 19.5 & 4.5 & 14.3 & 18.0 & 5.0 & n/a & n/a & 0.0 & 12.4 & \textbf{21.5} & 0.0 \\
\midrule
\multirow{5}{*}{GMAI} & InternVL-14B & 22.0 & 19.9 & 14.2 & n/a & n/a & 0.0 & 23.9 & 18.3 & 16.3 & 32.1 & \textbf{20.1} & 26.0 \\
 & Qwen3-VL-8B & 19.6 & \textbf{22.9} & 14.0 & 15.4 & 20.0 & 8.9 & 29.8 & \textbf{20.9} & 25.2 & n/a & n/a & 0.0 \\
 & Gemma-27B & 10.1 & \textbf{24.3} & 6.3 & 1.4 & \textbf{26.3} & 0.0 & n/a & n/a & 0.0 & 0.7 & 19.0 & 0.0 \\
 & LLaVA-13B & 0.5 & 10.5 & 0.0 & n/a & n/a & 0.0 & n/a & n/a & 0.0 & n/a & n/a & 0.0 \\
 & DeepSeek-VL2 & 0.4 & \textbf{27.7} & 0.0 & n/a & n/a & 0.0 & 0.4 & 9.8 & 0.0 & n/a & n/a & 0.0 \\
\midrule
\multirow{5}{*}{PATH-VQA} & InternVL-14B & 1.2 & \textbf{25.0} & 0.0 & n/a & n/a & 0.0 & n/a & n/a & 0.0 & 7.8 & 19.2 & 0.0 \\
 & Qwen3-VL-8B & 15.0 & \textbf{22.3} & 0.0 & 8.2 & 17.2 & 0.0 & 8.5 & 19.5 & 0.0 & n/a & n/a & 0.0 \\
 & Gemma-27B & 1.2 & 16.2 & 0.0 & 7.5 & 19.7 & 0.0 & n/a & n/a & 0.0 & n/a & n/a & 0.0 \\
 & LLaVA-13B & n/a & n/a & 0.0 & n/a & n/a & 0.0 & n/a & n/a & 0.0 & n/a & n/a & 0.0 \\
 & DeepSeek-VL2 & n/a & n/a & 0.0 & n/a & n/a & 0.0 & n/a & n/a & 0.0 & 1.2 & 13.9 & 0.0 \\
\bottomrule
\end{tabular}

\caption{Per-model operating points on the full calibration half at $\tau=0.20$, for the four estimators of Table~\ref{tab:budget}. Plug-in coverage and its realized test error, and certified coverage under Learn-then-Test with the Bentkus bound ($\delta=0.05$). Bold marks a realized error above 20\%; n/a marks a cell with no plug-in operating point.}
\label{tab:permodel_full}
\end{table*}

On the full calibration half, the plug-in threshold exceeds 20\% error on the test half in 14 of the 36 per-model cells where it yields an operating point, and the certified threshold in 1 of 20, at 20.90\% for BICR on Gemma-3-27B and GMAI-MMBench (Table~\ref{tab:permodel_full}). Across the label-budget grid at $\tau=0.20$, the plug-in threshold exceeds the tolerance in 169 of 340 operating points on SLAKE, 162 of 235 on GMAI-MMBench, and 78 of 203 on PATH-VQA (409 of 778 in total). The certified threshold does so in 0 of 131, 1 of 46, and 0 of 33. The single breach is the same model, probe, and dataset, at 20.26\% in one draw at 1{,}000 questions. At $\tau=0.10$, the counts are 350 of 514 for the plug-in rule and 0 of 34 for the certified rule. The pooled question-unit certificate exceeds its tolerance in none of its 54 certified cells. Over the 20 pooled certificates on the full calibration half and at the largest budgets, the realized question-averaged risk on the test half is at most 16.88\%, and the realized answer-level error at most 14.97\%. A guarantee at $\delta=0.05$ allows an occasional breach, and the two observed ones exceed the tolerance by less than one point.

\subsection{A fixed cut without labels}
\label{app:deployment:r0}

\begin{table*}[t]
\centering
\small
\begin{tabular}{@{}ll@{\hspace{7pt}}rr@{\hspace{7pt}}rr@{\hspace{7pt}}rr@{}}
\toprule
& & \multicolumn{2}{c}{BICR} & \multicolumn{2}{c}{II} & \multicolumn{2}{c}{PredEnt} \\
\cmidrule(lr){3-4} \cmidrule(lr){5-6} \cmidrule(lr){7-8}
Dataset & LVLM & Cov. \% & Err. \% & Cov. \% & Err. \% & Cov. \% & Err. \% \\
\midrule
\multirow{5}{*}{SLAKE} & InternVL-14B & 7.1 & 4.1 & 28.1 & 10.5 & 79.2 & \textbf{23.8} \\
 & Qwen3-VL-8B & 15.4 & 2.5 & 19.7 & 4.4 & 98.5 & \textbf{30.4} \\
 & Gemma-27B & 6.7 & 1.4 & 8.5 & 3.4 & 100.0 & \textbf{36.2} \\
 & LLaVA-13B & 8.1 & \textbf{31.8} & 12.1 & \textbf{38.6} & 69.8 & \textbf{47.5} \\
 & DeepSeek-VL2 & 1.4 & 0.0 & 0.8 & 0.0 & 15.0 & \textbf{29.9} \\
\midrule
\multirow{5}{*}{GMAI} & InternVL-14B & 28.1 & \textbf{21.4} & 26.0 & \textbf{31.0} & 99.8 & \textbf{39.0} \\
 & Qwen3-VL-8B & 2.6 & 8.6 & 6.9 & 12.3 & 100.0 & \textbf{45.8} \\
 & Gemma-27B & 6.3 & \textbf{20.9} & $<$0.1 & 0.0 & 100.0 & \textbf{51.7} \\
 & LLaVA-13B & 8.9 & \textbf{43.6} & 6.7 & \textbf{64.6} & 91.4 & \textbf{63.8} \\
 & DeepSeek-VL2 & $<$0.1 & 0.0 & 0.3 & \textbf{45.5} & 41.3 & \textbf{58.9} \\
\midrule
\multirow{5}{*}{PATH-VQA} & InternVL-14B & 4.2 & \textbf{34.6} & 10.8 & \textbf{33.6} & 85.0 & \textbf{51.4} \\
 & Qwen3-VL-8B & 5.0 & 14.9 & 9.6 & 17.5 & 98.1 & \textbf{51.2} \\
 & Gemma-27B & 0.6 & 10.5 & 1.2 & 13.2 & 100.0 & \textbf{53.8} \\
 & LLaVA-13B & 1.5 & \textbf{72.3} & 6.7 & \textbf{45.6} & 27.2 & \textbf{54.9} \\
 & DeepSeek-VL2 & 0.0 & n/a & 0.0 & n/a & 5.1 & \textbf{49.4} \\
\bottomrule
\end{tabular}

\caption{R0: a fixed cut at 0.8 on the raw score, with no labels. Coverage and realized error on the test half, per model. Bold marks an error above 20\%; n/a marks a cell that accepts nothing; $<$0.1 marks a nonzero coverage below 0.05\%. Mean token entropy (PredEnt) is shown for reference only; its scores are not probabilities, so a cut at 0.8 has no calibrated meaning for it.}
\label{tab:r0}
\end{table*}

Table~\ref{tab:r0} reports the fixed cut. For BICR and InternalInspector, it keeps the error at or below 20\% in 16 of the 28 cells where it accepts any answer and exceeds it in 12, with errors up to 72.3\%. A site without labels cannot tell in advance which case it is in.

\subsection{Per-model results at every budget}
\label{app:deployment:grid}

\begin{table*}[t]
\centering
\small
\begin{tabular}{@{}l@{\hspace{4pt}}l@{\hspace{5pt}}c@{\hspace{5pt}}c@{\hspace{5pt}}c@{\hspace{5pt}}c@{}}
\toprule
LVLM & Est. & $n{=}100$ & $n{=}250$ & $n{=}500$ & $n{=}1000$ \\
\midrule
InternVL-14B & BICR & 0.0 / 0.0 & 3.6$^{\dagger}$ / 11.3$^{\dagger}$ & 5.7$^{\dagger}$ / 43.0$^{\dagger}$ & 30.5 / 70.0 \\
 & II & 0.0 / 0.0 & 0.0 / 0.0 & 7.5$^{\dagger}$ / 8.0$^{\dagger}$ & 37.2 / 62.8 \\
\addlinespace[1pt]
Qwen3-VL-8B & BICR & 0.0 / 0.0 & 29.5$^{\dagger}$ / 0.0 & 48.7 / 5.1$^{\dagger}$ & 54.1 / 64.9 \\
 & II & 0.0 / 0.0 & 20.4$^{\dagger}$ / 0.0 & 41.7 / 0.0 & 54.2 / 60.1 \\
\addlinespace[1pt]
Gemma-27B & BICR & 0.0 / 0.0 & 6.4$^{\dagger}$ / 0.0 & 15.0$^{\dagger}$ / 18.9$^{\dagger}$ & 38.2 / 52.3 \\
 & II & 9.7$^{\dagger}$ / 0.0 & 19.6$^{\dagger}$ / 0.0 & 18.4$^{\dagger}$ / 0.0 & 49.6 / 35.8$^{\dagger}$ \\
\addlinespace[1pt]
LLaVA-13B & BICR & 0.0 / 0.0 & 0.0 / 0.0 & 0.0 / 0.0 & 0.0 / 0.0 \\
 & II & 0.0 / 0.0 & 0.0 / 0.0 & 0.0 / 0.0 & 0.0 / 0.0 \\
\addlinespace[1pt]
DeepSeek-VL2 & BICR & 0.0 / 0.0 & 0.0 / 0.0 & 0.0 / 4.2$^{\dagger}$ & 4.4 / 12.4$^{\dagger}$ \\
 & II & 0.0 / 0.0 & 0.0 / 0.0 & 0.0 / 0.0 & 5.6 / 0.0 \\
\midrule
\textit{Pooled} & BICR & 0.0 / 0.0 & 0.0 / 0.0 & 0.0 / 9.8$^{\dagger}$ & 6.8 / 37.6 \\
 & II & 0.0 / 0.0 & 0.0 / 0.0 & 0.0 / 0.0 & 7.7 / 18.4 \\
\bottomrule
\end{tabular}

\caption{SLAKE: certified coverage (\%) at $\tau=0.20$ for R1 / R2 at every budget, mean of five draws. Model rows are per model (Bentkus bound); the two pooled rows use the question-unit betting bound. A dagger marks a nonzero mean in which at least one draw certified nothing.}
\label{tab:grid_slake}
\end{table*}

\begin{table*}[t]
\centering
\small
\begin{tabular}{@{}l@{\hspace{4pt}}l@{\hspace{5pt}}c@{\hspace{5pt}}c@{\hspace{5pt}}c@{\hspace{5pt}}c@{\hspace{5pt}}c@{}}
\toprule
LVLM & Est. & $n{=}100$ & $n{=}250$ & $n{=}500$ & $n{=}1000$ & $n{=}2500$ \\
\midrule
InternVL-14B & BICR & 0.0 / 0.0 & 0.0 / 0.0 & 0.0 / 0.0 & 0.0 / 1.6$^{\dagger}$ & 4.0$^{\dagger}$ / 10.8$^{\dagger}$ \\
 & II & 0.0 / 0.0 & 0.0 / 0.0 & 0.0 / 0.0 & 0.0 / 0.0 & 0.0 / 20.0 \\
\addlinespace[1pt]
Qwen3-VL-8B & BICR & 0.0 / 0.0 & 0.0 / 0.0 & 0.0 / 0.0 & 0.0 / 4.4$^{\dagger}$ & 4.8$^{\dagger}$ / 19.8 \\
 & II & 0.0 / 0.0 & 0.0 / 0.0 & 0.0 / 0.0 & 0.0 / 2.6$^{\dagger}$ & 1.5$^{\dagger}$ / 19.6 \\
\addlinespace[1pt]
Gemma-27B & BICR & 0.0 / 0.0 & 0.0 / 0.0 & 0.0 / 0.0 & 1.1$^{\dagger}$ / 0.0 & 0.5$^{\dagger}$ / 2.5$^{\dagger}$ \\
 & II & 0.0 / 0.0 & 0.0 / 0.0 & 0.0 / 0.0 & 0.0 / 0.0 & 0.0 / 0.0 \\
\addlinespace[1pt]
LLaVA-13B & BICR & 0.0 / 0.0 & 0.0 / 0.0 & 0.0 / 0.0 & 0.0 / 0.0 & 0.0 / 0.0 \\
 & II & 0.0 / 0.0 & 0.0 / 0.0 & 0.0 / 0.0 & 0.0 / 0.0 & 0.0 / 0.0 \\
\addlinespace[1pt]
DeepSeek-VL2 & BICR & 0.0 / 0.0 & 0.0 / 0.0 & 0.0 / 0.0 & 0.0 / 0.0 & 0.0 / 0.0 \\
 & II & 0.0 / 0.0 & 0.0 / 0.0 & 0.0 / 0.0 & 0.0 / 0.0 & 0.0 / 0.0 \\
\midrule
\textit{Pooled} & BICR & 0.0 / 0.0 & 0.0 / 0.0 & 0.0 / 0.0 & 0.0 / 1.2$^{\dagger}$ & 0.4$^{\dagger}$ / 9.3 \\
 & II & 0.0 / 0.0 & 0.0 / 0.0 & 0.0 / 0.0 & 0.0 / 0.0 & 0.0 / 4.9$^{\dagger}$ \\
\bottomrule
\end{tabular}

\caption{GMAI-MMBench: as Table~\ref{tab:grid_slake}.}
\label{tab:grid_gmai}
\end{table*}

\begin{table*}[t]
\centering
\small
\begin{tabular}{@{}l@{\hspace{4pt}}l@{\hspace{5pt}}c@{\hspace{5pt}}c@{\hspace{5pt}}c@{\hspace{5pt}}c@{\hspace{5pt}}c@{}}
\toprule
LVLM & Est. & $n{=}100$ & $n{=}250$ & $n{=}500$ & $n{=}1000$ & $n{=}2500$ \\
\midrule
InternVL-14B & BICR & 0.0 / 0.0 & 0.0 / 0.0 & 0.0 / 0.0 & 0.0 / 0.0 & 0.0 / 16.1 \\
 & II & 0.0 / 0.0 & 0.0 / 0.0 & 0.0 / 0.0 & 0.0 / 0.0 & 0.0 / 10.9 \\
\addlinespace[1pt]
Qwen3-VL-8B & BICR & 0.0 / 0.0 & 0.0 / 0.0 & 0.0 / 0.0 & 0.0 / 0.0 & 0.0 / 28.9 \\
 & II & 0.0 / 0.0 & 0.0 / 0.0 & 0.0 / 0.0 & 0.0 / 0.0 & 0.0 / 30.7 \\
\addlinespace[1pt]
Gemma-27B & BICR & 0.0 / 0.0 & 0.0 / 0.0 & 0.0 / 0.0 & 0.0 / 0.0 & 0.0 / 14.1$^{\dagger}$ \\
 & II & 0.0 / 0.0 & 0.0 / 0.0 & 0.0 / 0.0 & 0.0 / 0.0 & 0.0 / 8.1$^{\dagger}$ \\
\addlinespace[1pt]
LLaVA-13B & BICR & 0.0 / 0.0 & 0.0 / 0.0 & 0.0 / 0.0 & 0.0 / 0.0 & 0.0 / 0.0 \\
 & II & 0.0 / 0.0 & 0.0 / 0.0 & 0.0 / 0.0 & 0.0 / 0.0 & 0.0 / 2.8$^{\dagger}$ \\
\addlinespace[1pt]
DeepSeek-VL2 & BICR & 0.0 / 0.0 & 0.0 / 0.0 & 0.0 / 0.0 & 0.0 / 0.0 & 0.0 / 0.0 \\
 & II & 0.0 / 0.0 & 0.0 / 0.0 & 0.0 / 0.0 & 0.0 / 0.0 & 0.0 / 3.7$^{\dagger}$ \\
\midrule
\textit{Pooled} & BICR & 0.0 / 0.0 & 0.0 / 0.0 & 0.0 / 0.0 & 0.0 / 0.0 & 0.0 / 11.4 \\
 & II & 0.0 / 0.0 & 0.0 / 0.0 & 0.0 / 0.0 & 0.0 / 0.0 & 0.0 / 9.9 \\
\bottomrule
\end{tabular}

\caption{PATH-VQA: as Table~\ref{tab:grid_slake}.}
\label{tab:grid_path}
\end{table*}

Tables~\ref{tab:grid_slake}, \ref{tab:grid_gmai}, and~\ref{tab:grid_path} give R1 and R2 for every model, probe, and budget. The counts of Section~\ref{sec:budget} are read from Table~\ref{tab:grid_slake}. On GMAI-MMBench, threshold-only calibration certifies more than adaptation in one cell, Gemma-3-27B with BICR at 1{,}000 questions, in one draw of five. On PATH-VQA, threshold-only calibration certifies nothing for any model at any budget.

Adapted BICR certifies only on calibration questions whose blank-image view was extracted, and blank views were extracted for the union of adaptation sets across draws and budgets. On GMAI-MMBench at 2{,}500 questions, R2 BICR therefore certifies on about 754 of roughly 1{,}250 calibration questions (60\%), and on PATH-VQA on about 1{,}197 (96\%); SLAKE is unaffected. Which questions are dropped depends on the random draws, not on their labels, and neither R1 nor the test half is affected. Fewer certifying questions can only loosen the bound, so R2 BICR's GMAI-MMBench certificates are conservative, and its comparison with R1 there uses about a third as many certifying labels as R1 does.

\subsection{Other adapted variants}
\label{app:deployment:variants}

\begin{table*}[t]
\centering
\small
\begin{tabular}{@{}l@{\hspace{3.5pt}}r@{\hspace{3.5pt}}r@{\hspace{3.5pt}}r@{\hspace{3.5pt}}r@{\hspace{6pt}}r@{\hspace{3.5pt}}r@{\hspace{3.5pt}}r@{\hspace{3.5pt}}r@{\hspace{3.5pt}}r@{\hspace{6pt}}r@{\hspace{3.5pt}}r@{\hspace{3.5pt}}r@{\hspace{3.5pt}}r@{\hspace{3.5pt}}r@{}}
\toprule
& \multicolumn{4}{c}{SLAKE} & \multicolumn{5}{c}{GMAI-MMBench} & \multicolumn{5}{c}{PATH-VQA} \\
\cmidrule(lr){2-5}\cmidrule(lr){6-10}\cmidrule(lr){11-15}
Variant & 100 & 250 & 500 & 1000 & 100 & 250 & 500 & 1000 & 2500 & 100 & 250 & 500 & 1000 & 2500 \\
\midrule
BICR, R2 & 0.0 & 0.0 & 9.8$^{\dagger}$ & 37.6 & 0.0 & 0.0 & 0.0 & 1.2$^{\dagger}$ & 9.3 & 0.0 & 0.0 & 0.0 & 0.0 & 11.4 \\
BICR, searched hyperparameters & 0.0 & 0.0 & 12.3$^{\dagger}$ & 38.4 & n/a & n/a & n/a & n/a & n/a & n/a & n/a & n/a & n/a & n/a \\
II, R2 & 0.0 & 0.0 & 0.0 & 18.4 & 0.0 & 0.0 & 0.0 & 0.0 & 4.9$^{\dagger}$ & 0.0 & 0.0 & 0.0 & 0.0 & 9.9 \\
II, common MLP head & 0.0 & 0.0 & 0.0 & 35.3 & 0.0 & 0.0 & 0.0 & 1.3$^{\dagger}$ & 11.1 & 0.0 & 0.0 & 0.0 & 0.0 & 12.2 \\
\bottomrule
\end{tabular}

\caption{Pooled certified coverage (\%) at $\tau=0.20$ for the two adapted probes and two variants, at every budget, mean of five draws. The searched-hyperparameter variant of BICR uses hyperparameters searched on the full SLAKE calibration half, so it uses more labels than its nominal budget and exists on SLAKE only. The common-head variant trains a shared MLP head on InternalInspector's features instead of InternalInspector's own encoder. A dagger marks a nonzero mean in which at least one draw certified nothing.}
\label{tab:r2variants}
\end{table*}

The R2 probes in the main text reuse the hyperparameters selected on GQA, so no labels are spent on a search. Table~\ref{tab:r2variants} reports two alternatives. BICR with hyperparameters searched on SLAKE certifies 38.4\% at 1{,}000 questions, against 37.6\% for R2, so the GQA hyperparameters cost little. A common MLP head on InternalInspector's features certifies more than InternalInspector's own encoder on all three datasets: 35.3\%, 11.1\%, and 12.2\% at the largest budgets, against 18.4\%, 4.9\%, and 9.9\%. The features support larger certificates than the original architecture reaches with these budgets.

\subsection{Recalibration instead of adaptation}
\label{app:deployment:recal}

\begin{table*}[t]
\centering
\small
\small
\begin{tabular}{@{}ll rrrrr@{}}
\toprule
Estimator & Arm & $n{=}100$ & $n{=}250$ & $n{=}500$ & $n{=}1000$ & $n{=}2500$ \\
\midrule
\multicolumn{7}{@{}l}{\textbf{SLAKE}} \\
BICR & R1 & 0.0 & 0.0 & 0.0 & 6.8 & n/a \\
 & R1 ($c$ only) & 0.0 & 0.0 & 0.0 & 1.2 & n/a \\
 & Platt & 0.0 & 0.0 & 0.0 & 15.0 & n/a \\
 & Isotonic & 0.0 & 0.0 & 0.0 & 11.8 & n/a \\
 & R2 & 0.0 & 0.0 & 9.8 & 37.6 & n/a \\
\midrule
II & R1 & 0.0 & 0.0 & 0.0 & 7.7 & n/a \\
 & R1 ($c$ only) & 0.0 & 0.0 & 0.0 & 0.0 & n/a \\
 & Platt & 0.0 & 0.0 & 3.6 & 15.7 & n/a \\
 & Isotonic & 0.0 & 0.0 & 0.0 & 11.7 & n/a \\
 & R2 & 0.0 & 0.0 & 0.0 & 18.4 & n/a \\
\midrule
PredEnt & R1 & 0.0 & 0.0 & 0.0 & 11.3 & n/a \\
 & R1 ($c$ only) & 0.0 & 0.0 & 0.0 & 2.1 & n/a \\
 & Platt & 0.0 & 0.0 & 0.0 & 12.7 & n/a \\
 & Isotonic & 0.0 & 0.0 & 0.0 & 11.7 & n/a \\
\midrule
\multicolumn{7}{@{}l}{\textbf{GMAI-MMBench}} \\
BICR & R1 & 0.0 & 0.0 & 0.0 & 0.0 & 0.4 \\
 & R1 ($c$ only) & 0.0 & 0.0 & 0.0 & 0.0 & 0.0 \\
 & Platt & 0.0 & 0.0 & 0.0 & 0.0 & 0.6 \\
 & Isotonic & 0.0 & 0.0 & 0.0 & 0.0 & 0.0 \\
 & R2 & 0.0 & 0.0 & 0.0 & 1.2 & 9.3 \\
\midrule
II & R1 & 0.0 & 0.0 & 0.0 & 0.0 & 0.0 \\
 & R1 ($c$ only) & 0.0 & 0.0 & 0.0 & 0.0 & 0.0 \\
 & Platt & 0.0 & 0.0 & 0.0 & 0.0 & 0.0 \\
 & Isotonic & 0.0 & 0.0 & 0.0 & 0.0 & 0.0 \\
 & R2 & 0.0 & 0.0 & 0.0 & 0.0 & 4.9 \\
\midrule
PredEnt & R1 & 0.0 & 0.0 & 0.0 & 0.6 & 5.0 \\
 & R1 ($c$ only) & 0.0 & 0.0 & 0.0 & 0.0 & 2.8 \\
 & Platt & 0.0 & 0.0 & 0.0 & 0.0 & 2.8 \\
 & Isotonic & 0.0 & 0.0 & 0.0 & 0.0 & 3.5 \\
\midrule
\multicolumn{7}{@{}l}{\textbf{PATH-VQA}} \\
BICR & R1 & 0.0 & 0.0 & 0.0 & 0.0 & 0.0 \\
 & R1 ($c$ only) & 0.0 & 0.0 & 0.0 & 0.0 & 0.0 \\
 & Platt & 0.0 & 0.0 & 0.0 & 0.0 & 0.0 \\
 & Isotonic & 0.0 & 0.0 & 0.0 & 0.0 & 0.0 \\
 & R2 & 0.0 & 0.0 & 0.0 & 0.0 & 11.4 \\
\midrule
II & R1 & 0.0 & 0.0 & 0.0 & 0.0 & 0.0 \\
 & R1 ($c$ only) & 0.0 & 0.0 & 0.0 & 0.0 & 0.0 \\
 & Platt & 0.0 & 0.0 & 0.0 & 0.0 & 0.0 \\
 & Isotonic & 0.0 & 0.0 & 0.0 & 0.0 & 0.0 \\
 & R2 & 0.0 & 0.0 & 0.0 & 0.0 & 9.9 \\
\midrule
PredEnt & R1 & 0.0 & 0.0 & 0.0 & 0.0 & 0.0 \\
 & R1 ($c$ only) & 0.0 & 0.0 & 0.0 & 0.0 & 0.0 \\
 & Platt & 0.0 & 0.0 & 0.0 & 0.0 & 0.0 \\
 & Isotonic & 0.0 & 0.0 & 0.0 & 0.0 & 0.0 \\
\bottomrule
\end{tabular}

\caption{Pooled certified coverage (\%) at $\tau=0.20$, mean of five draws, when R2's adaptation half is spent on a per-model Platt or isotonic map of the transferred score and the threshold is certified on R2's calibration half. R1 certifies the raw score on all $n$ labels; R1 ($c$ only) certifies it on the calibration half alone.}
\label{tab:recal}
\end{table*}

Instead of training a new probe on the adaptation half, a site could use it to recalibrate the transferred score. We fit a per-model Platt or isotonic map on R2's adaptation half and certify the threshold on its calibration half (Table~\ref{tab:recal}). Per model, a monotone map preserves the ranking, so it cannot certify more than R1, which uses twice the labels to certify. Pooled, the map puts the five models' scores on a common scale, which helps on SLAKE. At 1{,}000 questions, Platt scaling certifies 15.0\% for BICR and 15.7\% for InternalInspector, against 6.8\% and 7.7\% for R1 and 1.2\% and 0.0\% for the raw score certified on the same half. Recalibration never exceeds adapted BICR on any dataset or budget, and on GMAI-MMBench and PATH-VQA it certifies at most 2.8\%.

\subsection{Uneven splits of the label budget}
\label{app:deployment:split}

\begin{table*}[t]
\centering
\small
\small
\begin{tabular}{@{}llrrrrr@{}}
\toprule
& & \multicolumn{5}{c}{labeled questions $n$} \\
\cmidrule(lr){3-7}
Dataset \& probe & Allocation & 100 & 250 & 500 & 1000 & 2500 \\
\midrule
\multirow{4}{*}{SLAKE BICR} & R1 & 0.0 & 0.0 & 0.0 & 6.8 & n/a \\
 & R2 25/75 & 0.0 & 0.0 & 14.5$^{\dagger}$ & 32.1 & n/a \\
 & R2 50/50 & 0.0 & 0.0 & 9.8$^{\dagger}$ & 37.6 & n/a \\
 & R2 75/25 & 0.0 & 0.0 & 0.0 & 26.9$^{\dagger}$ & n/a \\
\midrule
\multirow{4}{*}{SLAKE II} & R1 & 0.0 & 0.0 & 0.0 & 7.7 & n/a \\
 & R2 25/75 & 0.0 & 0.0 & 0.0 & 15.0 & n/a \\
 & R2 50/50 & 0.0 & 0.0 & 0.0 & 18.4 & n/a \\
 & R2 75/25 & 0.0 & 0.0 & 0.0 & 9.0$^{\dagger}$ & n/a \\
\midrule
\multirow{4}{*}{GMAI-MMBench BICR} & R1 & 0.0 & 0.0 & 0.0 & 0.0 & 0.4$^{\dagger}$ \\
 & R2 25/75 & 0.0 & 0.0 & 0.0 & 0.8$^{\dagger}$ & 9.1 \\
 & R2 50/50 & 0.0 & 0.0 & 0.0 & 1.2$^{\dagger}$ & 9.3 \\
 & R2 75/25 & n/r & n/r & n/r & n/r & n/r \\
\midrule
\multirow{4}{*}{GMAI-MMBench II} & R1 & 0.0 & 0.0 & 0.0 & 0.0 & 0.0 \\
 & R2 25/75 & 0.0 & 0.0 & 0.0 & 0.0 & 0.0 \\
 & R2 50/50 & 0.0 & 0.0 & 0.0 & 0.0 & 4.9$^{\dagger}$ \\
 & R2 75/25 & n/r & n/r & n/r & n/r & n/r \\
\midrule
\multirow{4}{*}{PATH-VQA BICR} & R1 & 0.0 & 0.0 & 0.0 & 0.0 & 0.0 \\
 & R2 25/75 & 0.0 & 0.0 & 0.0 & 0.0 & 7.7 \\
 & R2 50/50 & 0.0 & 0.0 & 0.0 & 0.0 & 11.4 \\
 & R2 75/25 & 0.0 & 0.0 & 0.0 & 0.0 & 11.4 \\
\midrule
\multirow{4}{*}{PATH-VQA II} & R1 & 0.0 & 0.0 & 0.0 & 0.0 & 0.0 \\
 & R2 25/75 & 0.0 & 0.0 & 0.0 & 1.1$^{\dagger}$ & 3.2$^{\dagger}$ \\
 & R2 50/50 & 0.0 & 0.0 & 0.0 & 0.0 & 9.9 \\
 & R2 75/25 & 0.0 & 0.0 & 0.0 & 0.0 & 11.3$^{\dagger}$ \\
\bottomrule
\end{tabular}

\caption{Pooled certified coverage (\%) at $\tau=0.20$ under the question-unit betting bound, mean of five draws, when R2 spends 25\%, 50\%, or 75\% of the $n$ labels on adaptation and the rest on certification (written adaptation/certification); R1 spends all $n$ on certification. The 50/50 rows are the R2 values of Figure~\ref{fig:budget}. A dagger marks a nonzero mean in which at least one draw certified nothing; n/a marks a budget the dataset does not have; n/r marks a split not run. On GMAI-MMBench, BICR's certification sets are smaller than nominal (Appendix~\ref{app:deployment:grid}) and not in proportion to the split, so its rows compare splits only approximately.}
\label{tab:split}
\end{table*}

The main text splits the labels evenly. Table~\ref{tab:split} repeats R2 with a quarter or three quarters of the labels spent on adaptation, with everything else unchanged. At each dataset's largest budget, no uneven split certifies more than the even one beyond draw-to-draw variation: in nine of ten comparisons the mean favors the even split, in four of them in every draw, and the one exception, InternalInspector on PATH-VQA with three quarters spent on adaptation (11.3\% against 9.9\%), certifies in only four of five draws. At smaller budgets, spending less on adaptation certifies earlier. On SLAKE at 500 questions, BICR with a 25/75 split certifies 14.5\% in four draws, against 9.8\% in two for the even split, and on PATH-VQA at 1{,}000 questions a 25/75 split is the only allocation for which InternalInspector certifies at all (1.1\%, one draw). This matches the per-model pattern of Section~\ref{sec:budget}: when labels are scarce, the threshold needs more of them. The choice of split matters less than adapting at all: on SLAKE at 1{,}000 questions, every split certifies between 26.9\% and 37.6\% for BICR, against 6.8\% for R1. At $\tau=0.10$, the only pooled certificate under any split is BICR on SLAKE with a 25/75 split at 1{,}000 questions, which certifies 17.6\% in one draw of five. Initializing the adapted BICR from its GQA-trained weights instead of training from scratch changes the even-split certificates by at most 1.3 points and no certify-fraction. Across all splits, no certified threshold exceeds its tolerance on the test half.

\subsection{Strict tolerances}
\label{app:deployment:tight}

\begin{table*}[t]
\centering
\small
\begin{tabular}{@{}l@{\hspace{4pt}}l@{\hspace{6pt}}l@{\hspace{6pt}}r@{\hspace{4pt}}r@{\hspace{4pt}}r@{\hspace{4pt}}r@{}}
\toprule
& & & \multicolumn{4}{c}{Max budget} \\
\cmidrule(lr){4-7}
Dataset & LVLM & Unadap. best & R1 BICR & R2 BICR & R1 II & R2 II \\
\midrule
\multirow{5}{*}{SLAKE} & InternVL-14B & 26.5 (PredEnt) & 0.0 & 26.9$^{\dagger}$ & 0.0 & 14.9$^{\dagger}$ \\
 & Qwen3-VL-8B & 23.2 (BICR) & 21.6 & 8.4$^{\dagger}$ & 16.5 & 15.2$^{\dagger}$ \\
 & Gemma-27B & 0 & 0.0 & 0.0 & 0.0 & 3.3$^{\dagger}$ \\
 & LLaVA-13B & 0 & 0.0 & 0.0 & 0.0 & 0.0 \\
 & DeepSeek-VL2 & 0 & 0.0 & 0.0 & 0.0 & 0.0 \\
\midrule
\multirow{5}{*}{GMAI} & InternVL-14B & 0 & 0.0 & 0.0 & 0.0 & 0.0 \\
 & Qwen3-VL-8B & 10.7 (PredEnt) & 0.0 & 2.0$^{\dagger}$ & 0.0 & 4.4$^{\dagger}$ \\
 & Gemma-27B & 0 & 0.0 & 0.0 & 0.0 & 0.0 \\
 & LLaVA-13B & 0 & 0.0 & 0.0 & 0.0 & 0.0 \\
 & DeepSeek-VL2 & 0 & 0.0 & 0.0 & 0.0 & 0.0 \\
\midrule
\multirow{5}{*}{PATH-VQA} & InternVL-14B & 0 & 0.0 & 0.0 & 0.0 & 0.0 \\
 & Qwen3-VL-8B & 0 & 0.0 & 3.5$^{\dagger}$ & 0.0 & 2.2$^{\dagger}$ \\
 & Gemma-27B & 0 & 0.0 & 0.0 & 0.0 & 0.0 \\
 & LLaVA-13B & 0 & 0.0 & 0.0 & 0.0 & 0.0 \\
 & DeepSeek-VL2 & 0 & 0.0 & 0.0 & 0.0 & 0.0 \\
\bottomrule
\end{tabular}

\caption{Per-model certified coverage (\%) at $\tau=0.10$, with the same protocol and markers as Table~\ref{tab:budget}.}
\label{tab:tau10}
\end{table*}

At $\tau=0.10$, no pooled certificate exists in any of 77 cells, covering the full calibration half and every budget, with or without adaptation. Among the uneven splits of Appendix~\ref{app:deployment:split}, one cell certifies, in one draw of five. Per model, a few models certify, mostly on SLAKE (Table~\ref{tab:tau10}). At $\tau=0.05$, nothing is certified in any of 462 cells, per model or pooled. Both results also hold on the fixed GQA grid of Appendix~\ref{app:grid}.

\clearpage
\makeatletter\global\@colroom\@colht\global\vsize\@colht\makeatother
\section{What Adaptation Learns: Details}
\label{app:permodel}

\subsection{Question-text control for InternalInspector}
\label{app:permodel:text}

\begin{table*}[t]
\centering
\small
\small
\begin{tabular}{@{}l@{\hspace{6pt}}rrrr@{\hspace{7pt}}c@{\hspace{7pt}}rrrr@{}}
\toprule
& \multicolumn{4}{c}{AUROC} & Incr. & \multicolumn{4}{c}{Cert. \%} \\
\cmidrule(lr){2-5} \cmidrule(lr){7-10}
Dataset & Text & Text+R1 & R1 & R2 & R2 over Text & Text & Text+R1 & R1 & R2 \\
\midrule
SLAKE & 0.786 & 0.805 & 0.681 & 0.776 & $+0.029$ (21/25) & 13.3$^{\dagger}$ & 24.1 & 7.7 & 18.4 \\
GMAI-MMBench & 0.681 & 0.690 & 0.603 & 0.655 & $+0.019$ (17/25) & 0.0 & 0.0 & 0.0 & 4.9$^{\dagger}$ \\
PATH-VQA & 0.782 & 0.793 & 0.614 & 0.819 & $+0.049$ (25/25) & 0.0 & 0.0 & 0.0 & 9.9 \\
\bottomrule
\end{tabular}

\caption{Question-text control for InternalInspector, with the same protocol and columns as Table~\ref{tab:textcontrol}; the certified columns use the same question halves as Table~\ref{tab:textcontrol}. The text classifier is evaluated on the questions with InternalInspector scores, so its AUROC on GMAI-MMBench differs slightly from Table~\ref{tab:textcontrol}.}
\label{tab:textcontrol_ii}
\end{table*}

The incremental AUROC fits the text classifier with and without the probe score by 5-fold cross-fitting on the test half, with folds assigned by question, and reports the difference in out-of-fold AUROC with a question-clustered bootstrap interval over 25 model and draw tests. Table~\ref{tab:textcontrol_ii} repeats the control for InternalInspector. The text classifier out-ranks the transferred probe on all three datasets. Adding the transferred InternalInspector score to the text classifier out-ranks the adapted probe on SLAKE and GMAI-MMBench (0.805 against 0.776, 0.690 against 0.655) and certifies more on SLAKE (24.1\% against 18.4\%), so for this probe the gain from adaptation is small. On GMAI-MMBench, the text classifier is fit on more questions than the adapted probe because feature coverage shrinks the probe's calibration half; this favors the baseline, so the comparison with adapted BICR is conservative.

\subsection{Blank images}
\label{app:permodel:blank}

\begin{table}[ht]
\centering
\small
\begin{tabular}{@{}l@{\hspace{7pt}}r@{\hspace{7pt}}r@{\hspace{7pt}}r@{\hspace{7pt}}r@{}}
\toprule
Slice & Answers & Real \% & Blank \% & Lost \% \\
\midrule
\textit{SLAKE} \\
\quad closed & 4,180 & 67.3 & 40.3 & 54.2 \\
\quad open & 6,290 & 56.7 & 15.1 & 79.5 \\
\quad all & 10,470 & 60.9 & 25.1 & 68.3 \\
\addlinespace[2pt]
\textit{PATH-VQA} \\
\quad closed & 16,792 & 57.2 & 39.4 & 49.6 \\
\quad open & 13,755 & 17.9 & 3.5 & 86.6 \\
\quad all & 30,547 & 39.5 & 23.2 & 57.2 \\
\addlinespace[2pt]
\textit{GMAI-MMBench} \\
\quad all (multiple choice) & 25,000 & 47.4 & 26.0 & 60.6 \\
\midrule
\textbf{Overall} & 66,017 & 45.9 & 24.6 & \textbf{60.9} \\
\bottomrule
\end{tabular}

\caption{Accuracy with the real image and with a black image of the same size, pooled over the five models and graded by the same judge. Lost is the share of answers correct with the real image that are wrong with the blank one. GMAI-MMBench uses a 5{,}000-question subset and is entirely multiple choice.}
\label{tab:blank}
\end{table}

We replace each image with a black image of the same size, keep the question and prompt unchanged, and grade the new answers with the judge, which sees the real image. This covers 13{,}205 questions and 66{,}017 answers. Table~\ref{tab:blank} reports the results. The loss is largest on open-ended questions, where the answer cannot be guessed from a short list of options. On GMAI-MMBench, where the options give a measurable chance level of 23.8\%, blank-image accuracy stays close to it.

\subsection{Image replacement}
\label{app:permodel:grounding}

\begin{table*}[t]
\centering
\small
\begin{tabular}{@{}l@{\hspace{8pt}}l@{\hspace{8pt}}l@{}}
\toprule
Estimator & 4 LVLMs (main) & 5 LVLMs \\
\midrule
SeqLL & \textbf{$-5.58$ [$-6.02$, $-5.10$]} & \textbf{$-5.69$ [$-6.22$, $-5.12$]} \\
PredEnt & \textbf{$-3.39$ [$-3.66$, $-3.11$]} & \textbf{$-3.25$ [$-3.52$, $-2.97$]} \\
\addlinespace[2pt]
P(True) & \textbf{1.36 [0.62, 2.09]} & $-0.44$ [$-6.27$, 7.95] \\
SelfProbe & \textbf{$-2.32$ [$-3.13$, $-1.48$]} & \textbf{$-2.39$ [$-3.67$, $-1.25$]} \\
PromptEns & \textbf{$-0.70$ [$-1.02$, $-0.35$]} & \textbf{$-0.59$ [$-0.96$, $-0.25$]} \\
\addlinespace[2pt]
SAPLMA & \textbf{$-2.04$ [$-2.80$, $-1.26$]} & \textbf{$-5.85$ [$-7.98$, $-2.33$]} \\
InternalInspector & \textbf{$-2.02$ [$-2.49$, $-1.50$]} & \textbf{$-3.78$ [$-4.43$, $-3.07$]} \\
CCPS & $-0.78$ [$-1.77$, 0.24] & $-0.70$ [$-4.69$, 2.57] \\
BICR & $-0.72$ [$-1.65$, 0.20] & \textbf{$-4.39$ [$-5.98$, $-2.66$]} \\
\addlinespace[2pt]
adapted BICR (R2) & $-1.40$ (0+/3$-$) & 0.03 (0+/1$-$) \\
adapted InternalInspector (R2) & $-0.37$ (0+/2$-$) & $-0.29$ (0+/2$-$) \\
\midrule
\textit{warranted drop} & 1.71 & 3.21 \\
\bottomrule
\end{tabular}
\caption{Confidence drop (points) on image-invariant answers in GMAI-MMBench: mean confidence on the other answers minus that on image-invariant answers, computed within each model and averaged, with question-clustered 95\% bootstrap intervals. A positive drop means lower confidence on image-invariant answers; bold marks an interval that excludes zero. The main analysis averages the four models with at least 30 image-invariant answers; the second column adds Gemma-3-27B, which has 5. For the adapted probes, k+/k$-$ counts the five draws with a significantly positive or negative drop. The warranted drop is the accuracy gap between the two groups.}
\label{tab:replacement}
\end{table*}

For each of 5{,}000 GMAI-MMBench questions, we regenerate the answer with the image replaced, in turn, by three natural images from GQA, drawn deterministically per question from a pool that excludes the question's own image. An answer is image-invariant if its full text is unchanged under all three replacements. The number of image-invariant answers is 412 for InternVL3.5-14B, 475 for Qwen3-VL-8B, 1{,}106 for LLaVA-NeXT-13B, 199 for DeepSeek-VL2, and 5 for Gemma-3-27B. Since the drop is averaged with equal weight over models, Gemma-3-27B's term would rest on five answers. We therefore apply the paper's 30-answer floor and average over the other four models. Table~\ref{tab:replacement} reports both versions. Including Gemma-3-27B changes the verdict for two estimators: P(True) is no longer significant ($-0.44$), and BICR becomes significantly negative ($-4.39$). No trained probe lowers its confidence significantly in either version. The adapted probes are evaluated on the 2{,}493 replacement questions in the test half and never trained on them; no draw of either adapted probe shows a significantly positive drop in either version.

\noindent\textbf{SLAKE.} On SLAKE, only 258 of 10{,}470 answers (2.46\%) are image-invariant, and they are more accurate than the rest. The warranted drop is therefore negative ($-13.30$ points within models): a calibrated score should be \emph{more} confident on these answers. Lower confidence on image-invariant answers is then not the correct behavior, so the test cannot show whether a score registers image use, and we do not report it for SLAKE.

\clearpage
\section{Validation of the Correctness Judge}
\label{app:judge}

Every result in this study rests on binary correctness labels from one LLM judge, \texttt{gpt-5-mini}, which sees the image, the question, the gold answer, and the response. We check the judge in two ways. On closed items, a normalized string match against the gold answer provides an independent check over the whole population. On free-form items, where equivalence must be judged, a blind non-expert annotator, one of the authors, provides the reference.

\subsection{Agreement}
\label{app:judge:agreement}

\begin{table*}[t]
\centering
\small
\setlength{\tabcolsep}{2pt}

\caption{Validation of the automatic correctness judge on the two kinds of item it has to handle. Panel~(a) covers closed items, meaning all of GMAI-MMBench, which is entirely multiple choice, and the yes or no and short bounded answer questions of SLAKE and PATH-VQA. For these, correctness can be checked mechanically, so the judge is compared with a normalized string matcher over the entire closed population rather than a sample. Panel~(b) covers free text items, where semantic equivalence has to be judged rather than matched.}
\label{tab:supp:judge:overview}

\begin{tabular}{@{}l r r c c@{}}
\toprule
Slice & \multicolumn{1}{c}{$n$} & \multicolumn{1}{c}{Agreement} & 95\% CI & $\kappa$ \\
\midrule

\multicolumn{5}{@{}l}{\textbf{(a) Closed-form items, full population}} \\
\addlinespace[1pt]
All closed items & 128,613 & 97.91\% & --- & 0.958 \\
\quad GMAI-MMBench & 107,641 & 98.84\% & --- & 0.977 \\
\quad SLAKE & 4,180 & 90.77\% & --- & 0.801 \\
\quad PATH-VQA & 16,792 & 93.78\% & --- & 0.875 \\

\addlinespace[2pt]
Disagreements adjudicated & 150 & 91.3\% & [85.6, 95.3] & --- \\
Projected judge error & 128,613 & 0.18\% & [0.10, 0.30] & --- \\

\addlinespace[3pt]
\multicolumn{5}{@{}l}{\textbf{(b) Free-form items, human adjudication}} \\
\addlinespace[1pt]
All free-form items & 200 & 96.5\% & [92.9, 98.6] & 0.930 \\
\quad SLAKE & 100 & 97.0\% & [91.5, 99.4] & 0.940 \\
\quad PATH-VQA & 100 & 96.0\% & [90.1, 98.9] & 0.920 \\
\bottomrule
\end{tabular}

\vspace{4pt}

\begin{minipage}{\textwidth}
\footnotesize
\noindent
The string matcher is a reference, not ground truth: it fails on any answer its normalization cannot parse. A human therefore adjudicated a round-robin sample of 150 of the 2,682 cases where the two verdicts differ, and the following row projects that adjudication back onto the population; the implied judge accuracy on closed items is $\geq99.82\%$. The free-form panel reports agreement between the judge and a single annotator, blind to the judge's verdict and to any running agreement tally, on 200 items sampled 20 per LVLM per dataset, balanced on the judge's own label so both of its decisions are audited equally, and limited to one question per distinct image. \textbf{The population rows carry no interval because they are a census of the entire closed population, not a sample}; every interval shown is an exact Clopper--Pearson interval on a sampled quantity. These are two-rater agreement figures from one annotator; no third-party gold label exists for these items.
\end{minipage}

\end{table*}

On all 128{,}613 closed question and model pairs, the judge agrees with the string matcher on 97.9\% ($\kappa=0.958$; Table~\ref{tab:supp:judge:overview}). A disagreement is not necessarily a judge error, because the matcher fails on synonyms, option-letter conventions, and formatting. The annotator therefore adjudicated a sample of 150 of the 2{,}682 disagreements and found the judge correct in 91.3\%, including all 49 sampled from SLAKE. Projected to the full closed population, this puts the judge's error on closed items at 0.18\%, so its accuracy there is at least 99.8\%. On free-form items, the annotator labeled 200 responses: 100 per dataset and 20 per model, one per distinct image, balanced so that half carry a positive and half a negative judge label. Agreement is 96.5\% ($\kappa=0.930$) and is within one point of this on each dataset. These figures compare the judge with one non-expert annotator; no clinician reviewed the labels.

\subsection{Where free-form disagreements occur}
\label{app:judge:breakdown}

\begin{table}[t]
\centering
\footnotesize
\setlength{\tabcolsep}{2.5pt}
\caption{Where the free-form agreement of Table~\ref{tab:supp:judge:overview}(b) sits. Panel~(a) splits the 200 adjudicated items by LVLM, 40 each; panel~(b) splits them by the question subtype each benchmark provides. Neither the model nor the benchmark is a source of variation: every per-model figure lies within two items of the others.}
\label{tab:supp:judge:breakdown}
\begin{tabular}{@{}l r r c c@{}}
\toprule
 & \multicolumn{1}{c}{$n$} & \multicolumn{1}{c}{Agreement} & 95\% CI & $\kappa$ \\
\midrule
\multicolumn{5}{@{}l}{\textbf{(a) By LVLM}} \\
\addlinespace[1pt]
Qwen3-VL-8B & 40 & 95.0\% & [83.1, 99.4] & 0.900 \\
Gemma-3-27B & 40 & 95.0\% & [83.1, 99.4] & 0.900 \\
InternVL3.5-14B & 40 & 97.5\% & [86.8, 99.9] & 0.950 \\
LLaVA-NeXT-13B & 40 & 95.0\% & [83.1, 99.4] & 0.900 \\
DeepSeek-VL2 & 40 & 100.0\% & [91.2, 100.0] & 1.000 \\
\addlinespace[3pt]
\multicolumn{5}{@{}l}{\textbf{(b) By question subtype}} \\
\addlinespace[1pt]
PATH-VQA \textit{other} & 11 & 72.7\% & [39.0, 94.0] & 0.476 \\
SLAKE \textit{Organ} & 14 & 85.7\% & [57.2, 98.2] & 0.696 \\
SLAKE \textit{Position} & 9 & 88.9\% & [51.8, 99.7] & 0.780 \\
PATH-VQA \textit{why} & 11 & 90.9\% & [58.7, 99.8] & 0.814 \\
All 13 other subtypes & 155 & 100.0\% & [97.6, 100.0] & 1.000 \\
\bottomrule
\addlinespace[2pt]
\multicolumn{5}{@{}p{0.98\columnwidth}@{}}{\footnotesize All 7 free-form disagreements fall in the 4 small subtypes listed, while the remaining 13 subtypes (155 items) are in perfect agreement; the wide intervals in panel~(b) therefore reflect the small per-subtype $n$, not instability in the judge. Models appear in the fixed order used throughout this paper rather than sorted by agreement. Intervals are exact Clopper--Pearson; $\kappa$ is Cohen's $\kappa$ against the judge.} \\
\end{tabular}
\end{table}

Agreement ranges from 95\% to 100\% across the five models (Table~\ref{tab:supp:judge:breakdown}). All seven disagreements fall in four question subtypes with 9 to 14 items each; the other thirteen subtypes agree on every item.

\subsection{Direction of the residual error}
\label{app:judge:direction}

\begin{table*}[t]
\centering
\caption{What the judge gets wrong on free-form items, and in which direction. Because the adjudication set is balanced on the judge's own label, the quantities estimable without bias from it are those conditioned on that label, shown in panel~(a); the reverse conditioning (sensitivity and specificity with the annotator as reference) depends on the sampling design and is not reported. Panel~(b) reweights the two rates of panel~(a) by each benchmark's observed free-form judge accuracy, giving the accuracy a human grader would have recorded. \textbf{The judge's free-form errors are conservative}: it rejects answers a human credits far more often than the reverse, so the free-form accuracies reported in this paper are mild underestimates.}
\label{tab:supp:judge:direction}
\begin{threeparttable}
\footnotesize
\setlength{\tabcolsep}{3pt}
\begin{tabular}{@{}l r c r c@{}}
\toprule
\multicolumn{5}{@{}l}{\textbf{(a) Judge-conditional agreement}} \\
\cmidrule(r){1-5}
Slice & \multicolumn{1}{c}{P(human correct $\mid$ judge correct)} & 95\% CI & \multicolumn{1}{c}{P(human correct $\mid$ judge incorrect)} & 95\% CI \\
\midrule
All free-form items & 99.0\% (99/100) & [94.6, 100.0] & 6.0\% (6/100) & [2.2, 12.6] \\
\quad SLAKE & 100.0\% (50/50) & [92.9, 100.0] & 6.0\% (3/50) & [1.3, 16.5] \\
\quad PATH-VQA & 98.0\% (49/50) & [89.4, 99.9] & 6.0\% (3/50) & [1.3, 16.5] \\
\addlinespace[4pt]
\multicolumn{5}{@{}l}{\textbf{(b) Implied free-form accuracy}} \\
\cmidrule(r){1-5}
Free-form slice & \multicolumn{1}{c}{$n$} & Judge accuracy & \multicolumn{1}{c}{Human-implied accuracy} & Shift \\
\midrule
SLAKE, open answers & 6,290 & 56.66\% & 59.26\% & $+2.60$ pts \\
PATH-VQA, open answers & 13,755 & 17.95\% & 22.51\% & $+4.56$ pts \\
\bottomrule
\end{tabular}
\begin{tablenotes}[flushleft]
\footnotesize
\item The correction moves both benchmarks in the same direction, leaves the ordering between them unchanged, and does not touch any confidence-estimation result, all of which are computed from the judge's labels held fixed. The annotator was a single rater, blind to the judge's verdict and to any running agreement tally; these are two-rater agreement figures, and no third-party gold label exists for these items. Intervals are exact Clopper--Pearson.
\end{tablenotes}
\end{threeparttable}
\end{table*}

Because the free-form sample is balanced on the judge's label, the unbiased quantities are those conditioned on it (Table~\ref{tab:supp:judge:direction}). The annotator confirms 99.0\% of the judge's positive labels and overturns 94.0\% of its negative ones. Almost all of the judge's errors therefore reject answers the annotator credits. Reweighting by each dataset's free-form accuracy, an annotator would have recorded 2.6 points higher accuracy on SLAKE open answers and 4.6 points higher on PATH-VQA open answers. The labels are slightly strict. A strict label counts some correct accepted answers as errors, which makes certificates conservative.

\subsection{Other judges and prompts}
\label{app:judge:robustness}

\begin{table*}[t]
\centering
\caption{Cross-model comparison of the automatic correctness judge. Three LLM judges independently regrade the same 200 human-adjudicated free-form items under the paper's original rubric, each scored against the single human annotator. The right-hand columns give the direction of each judge's residual error against the human: how often it marks an answer correct, and whether its mistakes are rejections of answers the human credits (strict) or acceptances of answers the human rejects (lenient). The paper's judge sits at the human's own operating point and is not an outlier; the three disagree with the human in different directions, and their majority vote is not shown to improve on it.}
\label{tab:supp:judge:crossmodel}
\begin{threeparttable}
\scriptsize
\begin{tabular}{@{}l cc c rr@{}}
\toprule
 & \multicolumn{2}{c}{Agreement with human} & & \multicolumn{2}{c}{Errors vs human} \\
\cmidrule(lr){2-3}\cmidrule(lr){5-6}
Judge & \multicolumn{1}{c}{\%} & \multicolumn{1}{c}{95\% CI} & \multicolumn{1}{c}{$\kappa$} & \multicolumn{1}{c}{strict} & \multicolumn{1}{c}{lenient} \\
\midrule
GPT-5-mini \normalfont(the paper's judge) & 96.5 & [92.9, 98.6] & 0.930 & 6 & 1 \\
Claude Sonnet 5 & 97.0 & [93.6, 98.9] & 0.940 & 1 & 5 \\
Gemini 3 Flash & 92.0 & [87.3, 95.4] & 0.838 & 1 & 15 \\
\addlinespace[2pt]
Majority vote of the three & 97.5 & [94.3, 99.2] & --- & --- & --- \\
\midrule
\multicolumn{6}{@{}l}{\itshape\footnotesize Share of items marked correct: human 52.5\%, GPT-5-mini 50.0\%, Claude Sonnet 5 54.5\%, Gemini 3 Flash 59.5\%} \\
\bottomrule
\end{tabular}
\begin{tablenotes}[flushleft]
\footnotesize
\item Fleiss' $\kappa$ across the three model judges is 0.852; the human annotator is \emph{not} one of the three raters in that figure. Pairwise judge-to-judge agreement: GPT-5-mini vs Claude Sonnet 5 95.5\%; GPT-5-mini vs Gemini 3 Flash 89.5\%; Claude Sonnet 5 vs Gemini 3 Flash 93.0\%. The majority vote is the 2-of-3 label of the three model judges. It is nominally the highest entry in the table, but it exceeds the paper's judge by two items out of 200 with almost entirely overlapping intervals, so it is not evidence that an ensemble is better; it repairs five items the paper's judge got wrong and breaks three it got right. $\kappa$ is Cohen's $\kappa$ against the human. All figures are on the same 200 free-form items: 100 from each open-answer benchmark, 20 per LVLM per benchmark, one question per distinct image, and balanced on the original judge's own label so both of its decisions are audited equally. The reference throughout is a single annotator, blind to the judge's verdict and to any running agreement tally. Intervals are exact Clopper--Pearson. These are two-rater agreement figures against one annotator; no third-party gold label exists for these items, so a judge that disagrees with the annotator is differently calibrated rather than demonstrably wrong.
\end{tablenotes}
\end{threeparttable}
\end{table*}

\begin{table*}[t]
\centering
\caption{Prompt sensitivity of the automatic correctness judge. The judge model is held fixed at GPT-5-mini and only the grading rubric is varied, over the same 200 free-form items. The flip column reports how many labels move relative to the original rubric; the re-run row applies the original rubric a second time and is therefore the judge's own run-to-run variation, the floor against which the two rewordings should be read. Rewording moves no more labels than re-running the identical rubric does, so the labels are not an artefact of the specific prompt.}
\label{tab:supp:judge:prompt}
\begin{threeparttable}
\scriptsize
\begin{tabular}{@{}l cc cc@{}}
\toprule
 & \multicolumn{2}{c}{Labels flipped vs original} & \multicolumn{2}{c}{Agreement with human} \\
\cmidrule(lr){2-3}\cmidrule(lr){4-5}
Rubric & \multicolumn{1}{c}{\%} & \multicolumn{1}{c}{95\% CI} & \multicolumn{1}{c}{\%} & \multicolumn{1}{c}{95\% CI} \\
\midrule
Original rubric \normalfont(the paper's stored labels) & --- & --- & 96.5 & [92.9, 98.6] \\
Original rubric, re-run \normalfont(run-to-run variation) & 5.5 & [2.8, 9.6] & 93.0 & [88.5, 96.1] \\
Terse variant & 4.0 & [1.7, 7.7] & 94.0 & [89.8, 96.9] \\
Verbose variant & 6.0 & [3.1, 10.2] & 93.0 & [88.5, 96.1] \\
\bottomrule
\end{tabular}
\begin{tablenotes}[flushleft]
\footnotesize
\item The two rewordings are compared against the original rubric re-run in the same session, so the comparison is like-for-like. Re-running the identical rubric moves 5.5\% of labels on its own, more than the terse variant moves and within the interval of the verbose variant: at this sample size the effect of rewording the rubric is not separable from the judge's own run-to-run variation. The judge takes no seed, so that variation is irreducible. All three variants agree with the human within a point of one another. The stored labels agree with the human more closely than any fresh run of the same rubric, which is the same non-determinism seen from the other side. All figures are on the same 200 free-form items: 100 from each open-answer benchmark, 20 per LVLM per benchmark, one question per distinct image, and balanced on the original judge's own label so both of its decisions are audited equally. The reference throughout is a single annotator, blind to the judge's verdict and to any running agreement tally. Intervals are exact Clopper--Pearson. These are two-rater agreement figures against one annotator; no third-party gold label exists for these items, so a judge that disagrees with the annotator is differently calibrated rather than demonstrably wrong.
\end{tablenotes}
\end{threeparttable}
\end{table*}

On the same 200 free-form items and rubric, Claude Sonnet 5 agrees with the annotator on 97.0\% and Gemini 3 Flash on 92.0\% (Table~\ref{tab:supp:judge:crossmodel}). Fleiss' $\kappa$ across the three model judges is 0.852. The judges err in different directions. Against the annotator's 52.5\% positive rate, \texttt{gpt-5-mini} is slightly strict (6 of its 7 errors are rejections), Gemini 3 Flash is lenient (15 of its 16 errors are acceptances), and Claude Sonnet 5 is close to the annotator.

Holding the judge fixed and changing only the rubric, a terse and a verbose variant change 4.0\% and 6.0\% of labels (Table~\ref{tab:supp:judge:prompt}). Re-running the original rubric changes 5.5\% on its own, because the judge's outputs vary from run to run. Rewording therefore changes labels no more than repetition does, and all three rubrics agree with the annotator within one point of each other.

\clearpage
\section{Sensitivity to Unequal Error Costs}
\label{app:harm}

All main results count every error equally, whether the question asks for the imaging modality or for an abnormality. This appendix asks how much attainable coverage changes when errors on more consequential questions cost more. Certifying a harm-weighted risk would need a different bound, so we use the oracle yield of Appendix~\ref{app:metrics}: the largest coverage whose empirical risk is at most $\tau$, computed on all answers, calibration and test halves together. It shows the coverage a perfect choice of threshold could reach, not a certificate.

\subsection{Harm-weighted yield}
\label{app:harm:coverage}

\begin{table*}[t]
\centering
\caption{Harm-weighted oracle yield (\%) under three weighting schemes. Each answer in harm tier $t$ carries weight $w_t$. The accepted set is the largest tie-group-aligned prefix of the confidence-sorted answers whose harm-weighted error rate $\sum_{i \in S} w_i (1-y_i) / \sum_{i \in S} w_i$ is at most $\tau$ with at least 30 accepted answers, and HY@$\tau$ is its share of the weighted workload, $\sum_{i \in S} w_i / \sum_i w_i$. The schemes set $(w_1,w_2,w_3)$ to $(1,1,1)$, $(1,2,4)$, and $(1,3,9)$. The threshold is chosen on the same answers it is scored on, pooled over the five models and both halves, so these are ceilings rather than certified coverage. The uniform columns equal the unweighted oracle yield. Intervals are question-clustered bootstrap 95\% intervals (1{,}000 resamples). The ranking of estimators within each dataset is stable across schemes, apart from one adjacent exchange on SLAKE with overlapping intervals. On SLAKE at $\tau=0.20$, steep weighting roughly halves the yield of every estimator, and the uniform and steep intervals are disjoint.}
\label{tab:supp:harmweighted}
\resizebox{\textwidth}{!}{%
\begin{tabular}{@{}l rrr rrr@{}}
\toprule
 & \multicolumn{3}{c}{HY@20 [95\% CI]} & \multicolumn{3}{c}{HY@10 [95\% CI]} \\
\cmidrule(lr){2-4}\cmidrule(lr){5-7}
Estimator & \multicolumn{1}{c}{Uniform} & \multicolumn{1}{c}{Moderate} & \multicolumn{1}{c}{Steep} & \multicolumn{1}{c}{Uniform} & \multicolumn{1}{c}{Moderate} & \multicolumn{1}{c}{Steep} \\
\midrule
\multicolumn{7}{@{}l}{\textbf{GMAI-MMBench}} \\
\midrule
\addlinespace[1pt]
\multicolumn{7}{@{}l}{\itshape\footnotesize Inference-only baseline} \\
\addlinespace[1pt]
P(True) & 0.0 [0.0, 0.0] & 0.0 [0.0, 0.0] & 0.0 [0.0, 0.0] & 0.0 [0.0, 0.0] & 0.0 [0.0, 0.0] & 0.0 [0.0, 0.0] \\
\addlinespace[1pt]
\multicolumn{7}{@{}l}{\itshape\footnotesize Training-free logit baselines} \\
\addlinespace[1pt]
SeqLL$^{*}$ & 10.9 [10.3, 11.5] & 12.0 [11.4, 12.7] & 12.4 [11.7, 13.2] & 3.6 [2.7, 4.1] & 4.2 [3.6, 5.2] & 4.6 [3.7, 5.4] \\
PredEnt$^{*}$ & 10.9 [10.3, 11.5] & 12.1 [11.4, 12.8] & 12.4 [11.7, 13.2] & 3.6 [2.7, 4.1] & 4.2 [3.6, 5.2] & 4.5 [3.7, 5.4] \\
\addlinespace[1pt]
\multicolumn{7}{@{}l}{\itshape\footnotesize Trained probes} \\
\addlinespace[1pt]
InternalInspector & 0.0 [0.0, 0.2] & 0.0 [0.0, 0.2] & 0.0 [0.0, 0.2] & 0.0 [0.0, 0.0] & 0.0 [0.0, 0.0] & 0.0 [0.0, 0.0] \\
BICR & 4.5 [3.7, 5.4] & 6.9 [5.7, 7.9] & 7.7 [6.7, 8.6] & 0.9 [0.4, 1.4] & 1.4 [0.8, 1.9] & 1.4 [1.0, 2.2] \\
\midrule
\multicolumn{7}{@{}l}{\textbf{SLAKE}} \\
\midrule
\addlinespace[1pt]
\multicolumn{7}{@{}l}{\itshape\footnotesize Inference-only baseline} \\
\addlinespace[1pt]
P(True) & 29.5 [25.9, 32.6] & 20.9 [17.4, 24.5] & 16.3 [12.4, 19.7] & 9.5 [0.0, 15.3] & 0.0 [0.0, 8.1] & 0.0 [0.0, 0.0] \\
\addlinespace[1pt]
\multicolumn{7}{@{}l}{\itshape\footnotesize Training-free logit baselines} \\
\addlinespace[1pt]
SeqLL$^{*}$ & 25.1 [21.4, 27.9] & 15.6 [13.2, 19.2] & 12.1 [9.7, 14.8] & 11.1 [7.5, 14.0] & 5.5 [1.3, 8.0] & 2.5 [0.2, 5.2] \\
PredEnt$^{*}$ & 22.2 [19.9, 26.8] & 15.2 [12.8, 17.8] & 11.8 [9.4, 14.4] & 10.8 [7.9, 13.6] & 4.5 [1.3, 8.0] & 2.6 [0.2, 5.3] \\
\addlinespace[1pt]
\multicolumn{7}{@{}l}{\itshape\footnotesize Trained probes} \\
\addlinespace[1pt]
InternalInspector & 23.8 [19.7, 29.3] & 16.5 [13.4, 19.5] & 13.3 [10.5, 16.2] & 9.1 [6.6, 11.7] & 6.1 [4.4, 9.0] & 5.0 [2.7, 6.8] \\
BICR & 34.6 [30.2, 38.8] & 24.6 [20.9, 28.9] & 19.9 [15.7, 23.6] & 12.1 [7.3, 16.5] & 6.2 [4.3, 10.4] & 5.0 [2.6, 7.7] \\
\midrule
\multicolumn{7}{@{}l}{\textbf{PATH-VQA}} \\
\midrule
\addlinespace[1pt]
\multicolumn{7}{@{}l}{\itshape\footnotesize Inference-only baseline} \\
\addlinespace[1pt]
P(True) & 0.0 [0.0, 0.0] & 0.0 [0.0, 0.0] & 0.0 [0.0, 0.0] & 0.0 [0.0, 0.0] & 0.0 [0.0, 0.0] & 0.0 [0.0, 0.0] \\
\addlinespace[1pt]
\multicolumn{7}{@{}l}{\itshape\footnotesize Training-free logit baselines} \\
\addlinespace[1pt]
SeqLL$^{*}$ & 1.9 [1.0, 2.5] & 1.5 [0.9, 2.1] & 1.3 [0.9, 1.9] & 0.2 [0.0, 0.9] & 0.5 [0.0, 0.7] & 0.4 [0.0, 0.6] \\
PredEnt$^{*}$ & 1.9 [1.0, 2.4] & 1.5 [0.9, 2.1] & 1.4 [0.9, 1.9] & 0.2 [0.0, 0.9] & 0.5 [0.0, 0.7] & 0.4 [0.0, 0.6] \\
\addlinespace[1pt]
\multicolumn{7}{@{}l}{\itshape\footnotesize Trained probes} \\
\addlinespace[1pt]
InternalInspector & 1.3 [0.4, 2.2] & 1.2 [0.0, 2.3] & 1.1 [0.0, 2.4] & 0.0 [0.0, 0.5] & 0.0 [0.0, 0.3] & 0.0 [0.0, 0.3] \\
BICR & 0.0 [0.0, 1.0] & 0.0 [0.0, 0.9] & 0.0 [0.0, 0.9] & 0.0 [0.0, 0.0] & 0.0 [0.0, 0.0] & 0.0 [0.0, 0.0] \\
\bottomrule
\bottomrule
\end{tabular}%
}

\vspace{3pt}
\parbox{\textwidth}{%
\footnotesize
\raggedright
Intervals are 95\% clustered-bootstrap percentile intervals over 1{,}000 resamples with the cluster taken to be the question, so a question's five LVLM rows always resample together; results are pooled over the five LVLMs. Within a dataset a single resample stream is shared across schemes and estimators, so the columns are paired by construction. Harm tiers for the 41 question types were drafted with an LLM and reviewed by the authors, are not validated by a clinical expert, and serve only as the axis of this sensitivity analysis. $^{*}$~Training-free logit method; its absolute scale is a truncated-softmax approximation rather than a probability.
}
\end{table*}

Each question type was assigned one of three ordinal harm tiers (low, moderate, high), drafted with an LLM and reviewed by the authors. An answer of tier $k$ receives weight $w_k$ in both the coverage and the error rate, and the harm-weighted yield HY@$\tau$ is the largest share of the weighted workload whose weighted error rate is at most $\tau$. Table~\ref{tab:supp:harmweighted} compares equal weights, moderate weights $(1,2,4)$, and steep weights $(1,3,9)$. The ranking of estimators within each dataset is almost unchanged across the three schemes. The amount of coverage is not: on SLAKE at $\tau=0.20$, steep weights roughly halve the yield of every estimator, and the equal-weight and steep intervals do not overlap. On GMAI-MMBench, the yield instead rises slightly as the weights steepen.

\subsection{Why the schemes move differently}
\label{app:harm:mechanism}

\begin{table*}[t]
\centering
\caption{Where accepted answers fall, and accuracy by harm tier. Panel (a) gives the harm-tier composition of the answers each estimator accepts at $\tau=0.20$ under equal weights (oracle threshold, all answers), beside the composition of the dataset. On SLAKE the accepted set is enriched in tier~1 and depleted in tier~3; for BICR, 26.4\% against a 19.5\% dataset share and 5.3\% against 14.9\%. On GMAI-MMBench the logit baselines accept a set slightly enriched in tier~3 (60.3\% against 56.5\%). Panel (b) gives the models' accuracy by tier. Accuracy falls with harm on SLAKE and PATH-VQA but rises on GMAI-MMBench, where the high-harm tasks (disease diagnosis, severity grading, microorganism recognition) are among those the models answer best and the low-harm tasks (attribute recognition, image quality grading) among those they answer worst. Upweighting harm on GMAI-MMBench therefore upweights accurate work, so its harm-weighted yield rises as the weights steepen. All values are percentages.}
\label{tab:supp:harmwhere}
\begin{threeparttable}
\footnotesize
\begin{tabular}{@{}l r rrr@{}}
\toprule
\multicolumn{5}{@{}l}{\textbf{(a) Harm tier composition of the accepted set at $\tau=0.20$}} \\
\addlinespace[2pt]
Set & \multicolumn{1}{c}{Answers accepted} & \multicolumn{1}{c}{Tier 1} & \multicolumn{1}{c}{Tier 2} & \multicolumn{1}{c}{Tier 3} \\
\midrule
\multicolumn{5}{@{}l}{\textbf{GMAI-MMBench}} \\
\midrule
Dataset (all answers) & 107,641 & 10.0 & 33.5 & 56.5 \\
P(True) accepted & 0 & \textit{n/a} & \textit{n/a} & \textit{n/a} \\
SeqLL$^{*}$ accepted & 11,690 & 6.3 & 33.4 & 60.3 \\
PredEnt$^{*}$ accepted & 11,706 & 6.3 & 33.4 & 60.3 \\
InternalInspector accepted & 0 & \textit{n/a} & \textit{n/a} & \textit{n/a} \\
BICR accepted & 4,872 & 8.2 & 39.9 & 51.9 \\
\midrule
\multicolumn{5}{@{}l}{\textbf{SLAKE}} \\
\midrule
Dataset (all answers) & 10,470 & 19.5 & 65.6 & 14.9 \\
P(True) accepted & 3,093 & 25.2 & 67.5 & 7.4 \\
SeqLL$^{*}$ accepted & 2,630 & 30.2 & 62.8 & 7.0 \\
PredEnt$^{*}$ accepted & 2,327 & 31.6 & 62.4 & 6.0 \\
InternalInspector accepted & 2,488 & 23.1 & 69.1 & 7.8 \\
BICR accepted & 3,623 & 26.4 & 68.3 & 5.3 \\
\midrule
\multicolumn{5}{@{}l}{\textbf{PATH-VQA}} \\
\midrule
Dataset (all answers) & 30,547 & 0.0 & 39.0 & 61.0 \\
P(True) accepted & 0 & \textit{n/a} & \textit{n/a} & \textit{n/a} \\
SeqLL$^{*}$ accepted & 587 & 0.0 & 78.5 & 21.5 \\
PredEnt$^{*}$ accepted & 595 & 0.0 & 78.7 & 21.3 \\
InternalInspector accepted & 400 & 0.0 & 49.2 & 50.7 \\
BICR accepted & 0 & \textit{n/a} & \textit{n/a} & \textit{n/a} \\
\midrule
\addlinespace[3pt]
\multicolumn{5}{@{}l}{\textbf{(b) Base VQA accuracy by harm tier}} \\
\addlinespace[2pt]
Dataset & \multicolumn{1}{c}{Question types} & \multicolumn{1}{c}{Tier 1} & \multicolumn{1}{c}{Tier 2} & \multicolumn{1}{c}{Tier 3} \\
\midrule
GMAI-MMBench & 18 & 38.1 & 44.6 & 50.4 \\
SLAKE & 10 & 79.3 & 59.4 & 43.3 \\
PATH-VQA & 13 & \textit{n/a} & 48.5 & 33.7 \\
\bottomrule
\end{tabular}
\begin{tablenotes}[flushleft]
\footnotesize
\item Panel (a) is computed under equal cost, the assumption in force elsewhere in the paper, so it describes the accepted set. \textit{n/a} marks an estimator that accepts nothing at $\tau=0.20$ on that dataset, for which no composition is defined; the zero is retained because accepting nothing is itself the result. PATH-VQA has no tier-1 question type, so its tier-1 accuracy is likewise undefined. Accuracies in panel (b) are pooled over the five LVLMs and are the models' own VQA accuracy, not an estimator quantity. Harm tiers for the 41 question types were drafted with an LLM and reviewed by the authors, are not validated by a clinical expert, and serve only as the axis of this sensitivity analysis rather than as a clinical instrument. $^{*}$~Training-free logit method.
\end{tablenotes}
\end{threeparttable}
\end{table*}

Table~\ref{tab:supp:harmwhere} explains the difference. On SLAKE, the answers accepted at $\tau=0.20$ are mostly low-harm, and high-harm questions make up between a third and a half of their share of the dataset, depending on the estimator (panel a). Upweighting high-harm questions therefore enlarges the part of the workload that is not accepted. Panel (b) shows why GMAI-MMBench behaves the other way. On SLAKE and PATH-VQA, accuracy falls from low-harm to high-harm questions. On GMAI-MMBench it rises: disease diagnosis and severity grading are among the tasks the models answer best, while attribute recognition and image-quality grading are among the worst. Whether the questions a model answers least reliably are also the least consequential depends on the dataset and cannot be assumed.

\subsection{Scope}
\label{app:harm:scope}

The tiers are an ordinal judgment at the level of question types, drafted with an LLM, reviewed by the authors, and not validated by a clinical expert, and the weights are illustrative. The analysis is a sensitivity check and supports no claim in the main text beyond the limitation that all errors count equally there.

\clearpage
\clearpage
\section{Inference Cost}
\label{app:cost}

Tables~\ref{tab:supp:latency} and~\ref{tab:supp:memory} report the latency and peak GPU memory each estimator adds on top of generation, for all nine estimators and five models, measured on a single NVIDIA H200. The two logit baselines are computed from logits that generation already produces, so they add under a millisecond and almost no memory. The trained probes add one forward pass over stored internal states: BICR and InternalInspector add 177 to 625\,ms on the four models run in float32, against 248 to 2{,}121\,ms for generation. P(True) and Self-Probing each cost about one more forward pass of the model. Prompt Ensemble runs the model once per paraphrase and costs about ten times generation, up to 25\,s per question on the slowest model. DeepSeek-VL2 ran in bfloat16 in the environment its code requires, so its absolute figures are not directly comparable with the other four. The memory table reports the transient peak of each estimator's computation and excludes the stored weights of a trained head (up to 10.5\,MiB for BICR and 43.2\,MiB for InternalInspector). SAPLMA's peak memory was not measured. Adapting a probe under R2 does not change these costs, because the adapted probe has the same architecture as the transferred one.

\paragraph{Total compute and software.} The five VLMs have 8B (Qwen3-VL-8B), 13B (LLaVA-NeXT-13B), 14B (InternVL3.5-14B), and 27B (Gemma-3-27B) parameters; DeepSeek-VL2 is a mixture-of-experts model with about 27B parameters, of which 4.5B are active per token. All generation, feature extraction, probe training and certification used about 217 GPU-hours on NVIDIA H200 GPUs (an estimate), plus about \$41 of API calls for the correctness judges. Generation used Hugging Face Transformers 4.57.3; DeepSeek-VL2 ran in a separate environment with Transformers 4.38.2, as its released code requires. Exact versions for every environment are pinned in the code supplement.

\begin{table*}[t]
\centering
\caption{Inference cost of the confidence layer: median added latency per case, in milliseconds, for every estimator on every LVLM, beside the number of trained parameters each carries. Latency is the time the estimator adds on top of generation, whose answer already exists when the estimator runs; the generation row is the shared reference. The training-free logit baselines are effectively free, the trained probes add a single forward pass well below generation on the larger models, and Prompt Ensemble, which re-runs the model once per prompt, costs roughly an order of magnitude more than generation throughout.}
\label{tab:supp:latency}
\begin{threeparttable}
\small
\begin{tabular}{@{}l rrrrr r@{}}
\toprule
Estimator & \multicolumn{1}{c}{Qwen-8B} & \multicolumn{1}{c}{Gemma-27B} & \multicolumn{1}{c}{InternVL-14B} & \multicolumn{1}{c}{LLaVA-13B} & \multicolumn{1}{c}{DeepSeek} & \multicolumn{1}{c}{Trained params} \\
\midrule
\textit{Generation (reference)} & 247.5 & 871.1 & 2{,}120.6 & 2{,}303.2 & 666.9 & none \\
\addlinespace[1pt]
\multicolumn{7}{@{}l}{\itshape\footnotesize Training-free logit baselines} \\
\addlinespace[1pt]
SeqLL$^{*}$ & 0.4 & 0.3 & 0.4 & 0.5 & 0.3 & none \\
PredEnt$^{*}$ & 0.3 & 0.3 & 0.3 & 0.3 & 0.2 & none \\
\addlinespace[1pt]
\multicolumn{7}{@{}l}{\itshape\footnotesize Trained probes} \\
\addlinespace[1pt]
SAPLMA & 0.1 & 0.1 & 0.1 & 0.1 & 0.1 & 0.70--1.42\,M \\
BICR & 176.4 & 624.7 & 276.6 & 1{,}999.9 & 274.3 & 0.53--2.75\,M \\
InternalInspector & 176.4 & 624.7 & 276.6 & 1{,}999.9 & 274.3 & 11.33\,M \\
CCPS & 902.0 & 3{,}132.4 & 1{,}384.8 & 10{,}013.1 & 1{,}416.6 & 21{,}778 \\
\addlinespace[1pt]
\multicolumn{7}{@{}l}{\itshape\footnotesize Inference-only baselines} \\
\addlinespace[1pt]
P(True) & 198.9 & 674.4 & 320.5 & 2{,}007.0 & 318.7 & none \\
Self-Probing & 473.3 & 1{,}121.1 & 592.9 & 2{,}236.7 & 649.4 & none \\
Prompt Ensemble & 2{,}639.2 & 9{,}241.2 & 23{,}024.8 & 25{,}268.4 & 7{,}335.0 & none \\
\bottomrule
\end{tabular}
\begin{tablenotes}[flushleft]
\footnotesize
\item All five LVLMs were timed on the same single NVIDIA H200 (139.8\,GiB) with no multi-GPU device map, so absolute milliseconds are comparable across columns as well as across rows. Protocol: SLAKE, batch size 1, greedy decoding, \texttt{max\_new\_tokens}=64 (Self-Probing 128), 11 prompts for Prompt Ensemble, 5 perturbation steps for CCPS, top-$k$=100, 250 timed samples per model after warmup, every call CUDA-synchronised with its own peak-memory reset. Four models ran in float32 under torch 2.9.0; DeepSeek-VL2 ran in bfloat16 under torch 2.9.1 with transformers 4.38.2, the pinned environment its vendored modelling code requires, so its absolute milliseconds carry an interpreter difference the other four do not. BICR and InternalInspector are timed identically because both are composed from the same teacher-forced replay primitive and read the same stored states; they differ only in the size of the head trained on top, which the parameter column reports. Parameter counts are ranges where the head is sized to the LVLM's hidden dimension and a single figure where it is fixed; CCPS's classifier is 21{,}778 parameters on every model. One accounting inconsistency is inherited from the original benchmark and is flagged rather than silently repaired: BICR is composed as replay plus probe forward while SAPLMA is composed as probe forward alone, on the grounds that generation has already produced SAPLMA's hidden states. Both probes read the same stored representation, so the two conventions cannot both be right, and SAPLMA's latency here is a lower bound under the alternative convention. $^{*}$~Training-free logit method; its absolute scale is a truncated-softmax approximation rather than a probability, so any fixed-threshold entry for it is reported for completeness and is not a calibration claim.
\end{tablenotes}
\end{threeparttable}
\end{table*}
\begin{table*}[t]
\centering
\caption{Inference cost of the confidence layer: peak GPU memory overhead per case, in MiB, measured above the resident model for every estimator on every LVLM. Each call is wrapped in its own peak-memory reset and the delta over the loaded model is recorded. The logit baselines add a negligible transient, while the trained probes and inference-only methods add a working set that can approach or exceed generation's own peak on the larger models.}
\label{tab:supp:memory}
\begin{threeparttable}
\small
\begin{tabular}{@{}l rrrrr@{}}
\toprule
Estimator & \multicolumn{1}{c}{Qwen-8B} & \multicolumn{1}{c}{Gemma-27B} & \multicolumn{1}{c}{InternVL-14B} & \multicolumn{1}{c}{LLaVA-13B} & \multicolumn{1}{c}{DeepSeek} \\
\midrule
\textit{Generation (reference)} & 342.2 & 2{,}162.0 & 451.9 & 9{,}817.2 & 972.8 \\
\addlinespace[1pt]
\multicolumn{6}{@{}l}{\itshape\footnotesize Training-free logit baselines} \\
\addlinespace[1pt]
SeqLL$^{*}$ & $<$0.1 & $<$0.1 & $<$0.1 & $<$0.1 & $<$0.1 \\
PredEnt$^{*}$ & $<$0.1 & $<$0.1 & $<$0.1 & $<$0.1 & $<$0.1 \\
\addlinespace[1pt]
\multicolumn{6}{@{}l}{\itshape\footnotesize Trained probes} \\
\addlinespace[1pt]
SAPLMA & n/m$^{\S}$ & n/m$^{\S}$ & n/m$^{\S}$ & n/m$^{\S}$ & n/m$^{\S}$ \\
BICR & 444.6 & 2{,}162.1 & 617.5 & 9{,}862.3 & 976.3 \\
InternalInspector & 444.6 & 2{,}162.1 & 617.5 & 9{,}862.3 & 976.3 \\
CCPS & 878.7 & 2{,}848.5 & 1{,}229.7 & 17{,}300.4 & 1{,}692.7 \\
\addlinespace[1pt]
\multicolumn{6}{@{}l}{\itshape\footnotesize Inference-only baselines} \\
\addlinespace[1pt]
P(True) & 292.8 & 2{,}162.7 & 373.8 & 7{,}770.5 & 852.3 \\
Self-Probing & 209.6 & 2{,}163.2 & 281.9 & 7{,}863.9 & 1{,}026.9 \\
Prompt Ensemble & 184.3 & 2{,}162.0 & 207.6 & 7{,}621.1 & 972.7 \\
\bottomrule
\end{tabular}
\begin{tablenotes}[flushleft]
\footnotesize
\item All five LVLMs were measured on the same single NVIDIA H200 (139.8\,GiB) with no multi-GPU device map, so absolute MiB are comparable across columns as well as across rows. Each figure is the peak GPU memory the estimator's forward work adds above the already-resident model, recorded under the same protocol as the latency benchmark: SLAKE, batch size 1, greedy decoding, \texttt{max\_new\_tokens}=64 (Self-Probing 128), 11 prompts for Prompt Ensemble, 5 perturbation steps for CCPS, top-$k$=100, every call wrapped in its own peak-memory reset. Four models ran in float32; DeepSeek-VL2 ran in bfloat16 in the pinned environment its vendored modelling code requires, so its column is not directly comparable in absolute terms to the other four. BICR and InternalInspector coincide because both are composed from the same replay primitive and read the same stored states. The overhead reported here is the transient peak of the estimator's forward work; it excludes the resident footprint of a trained head, which is 2.03--10.50\,MiB for BICR, 43.21\,MiB for InternalInspector, 4.16--5.41\,MiB for SAPLMA and 0.08\,MiB for CCPS, so the shared BICR/InternalInspector row should not be read as equal total cost. The two logit baselines are arithmetic over the top-$k$ array generation already carries; their measured overhead spans 0.003--0.030\,MiB. $^{\S}$~n/m, not measured: the probe-forward was not wrapped for peak-memory capture, so no figure is available. It is reported as unmeasured rather than as zero. $^{*}$~Training-free logit method; its absolute scale is a truncated-softmax approximation rather than a probability, so any fixed-threshold entry for it is reported for completeness and is not a calibration claim.
\end{tablenotes}
\end{threeparttable}
\end{table*}

\clearpage

\clearpage
\section{Dataset Details}
\label{app:data}

\paragraph{Licenses.} SLAKE is distributed under CC-BY-4.0, PATH-VQA under MIT, and GMAI-MMBench under Apache-2.0; underlying image rights remain with the original sources. GQA, used only to train the transferred estimators, is used under its published terms. Models were used under their published licenses. We collect no new patient data.

\paragraph{Supplementary Data.} The supplementary data archive contains a small example subset drawn from the three public test sets: 150 images (50 per dataset), the closed-form questions about each image, the gold answers, each VLM's answer and correctness label, and the nine transferred confidence scores. These samples are provided solely for scientific peer review and verification, not as a standalone dataset release. The full datasets must be obtained from their original sources under the applicable terms. The original benchmark sources and their respective licenses are acknowledged. We removed file metadata (EXIF, text chunks, ICC profiles) from the included images to reduce potential information exposure.

\paragraph{Language.} SLAKE is bilingual, and we evaluate both its English and Chinese questions (about 51\% and 49\% of its 2,094 questions); PATH-VQA and GMAI-MMBench are English only. All reported SLAKE results are computed over both languages. For readability, the example subset in the supplementary material shows only English SLAKE questions. SLAKE covers radiology images (CT, MRI, X-ray), PATH-VQA covers pathology images, and GMAI-MMBench covers clinical tasks across multiple imaging modalities.

\clearpage

\end{document}